\documentclass[[10pt,twocolumn,twoside]{IEEEtran}

\usepackage{amsmath,amsfonts}
\usepackage{algorithmic}
\usepackage{algorithm}
\usepackage{array}
\usepackage[caption=false,font=normalsize,labelfont=sf,textfont=sf]{subfig}
\usepackage{textcomp}
\usepackage{stfloats}
\usepackage{url}
\usepackage{verbatim}
\usepackage{graphicx}
\usepackage{cite}
\usepackage{booktabs} %调整表格线与上下内容的间隔
\usepackage{times}
\usepackage{epsfig}
\usepackage{arydshln}
\usepackage{amssymb}
\usepackage{multirow}
\usepackage{stackengine}

\usepackage[pagebackref=true,breaklinks=true,letterpaper=true,colorlinks,bookmarks=false]{hyperref}

\begin{document}

\title{Learning to Pan-sharpening with Memories of Spatial Details}

\author{Maoxun Yuan, Tianyi Zhao, Bo Li and Xingxing Wei$^\ddagger$

\thanks{Maoxun Yuan and Tianyi Zhao are co-first authors. $\ddagger$ indicates the corresponding author.}

\thanks{Maoxun Yuan and Bo Li are with the Beijing Key Laboratory of Digital Media, School of Computer Science and Engineering, Beihang University, Beijing
100191, China (email: yuanmaoxun@buaa.edu.cn; boli@buaa.edu.cn).}% <-this % stops a space
% \thanks{Bo Li is with with Institute of Artificial Intelligence, Beihang University, Beijing
% 100191, China.}% <-this % stops a space
\thanks{Xingxing Wei and Tianyi Zhao are with Institute of Artificial Intelligence, Beihang University, Beijing, 100191, China (email: xxwei@buaa.edu.cn; Ty\_Zhao@buaa.edu.cn). }
}

\markboth{Journal of \LaTeX\ Class Files,~Vol.~18, No.~9, September~2020}%
{Yuan \MakeLowercase{\textit{et al.}}: Memories of Spatial Details}

% \IEEEpubid{0000--0000/00\$00.00~\copyright~2023 IEEE}


\maketitle

\begin{abstract}
Pan-sharpening, as one of the most commonly used techniques in remote sensing systems, aims to inject spatial details from panchromatic images into multispectral images (MS) to obtain high-resolution multispectral images. Since deep learning has received widespread attention because of its powerful fitting ability and efficient feature extraction, a variety of pan-sharpening methods have been proposed to achieve remarkable performance. However, current pan-sharpening methods usually require the paired panchromatic (PAN) and MS images as input, which limits their usage in some scenarios. To address this issue, in this paper we observe that the spatial details from PAN images are mainly high-frequency cues, i.e., the edges reflect the contour of input PAN images. This motivates us to develop a PAN-agnostic representation to store some base edges, so as to compose the contour for the corresponding PAN image via them. As a result, we can perform the pan-sharpening task with only the MS image when inference. To this end, a memory-based network is adapted to extract and memorize the spatial details during the training phase and is used to replace the process of obtaining spatial information from PAN images when inference, which is called Memory-based Spatial Details Network (MSDN).  Finally, we integrate the proposed MSDN module into the existing deep learning-based pan-sharpening methods to achieve an end-to-end pan-sharpening network. With extensive experiments on the Gaofen1 and WorldView-4 satellites, we verify that our method constructs good spatial details without PAN images and achieves the best performance.
The code is available at \href{https://github.com/Zhao-Tian-yi/Learning-to-Pan-sharpening-with-Memories-of-Spatial-Details.git}{https://github.com/Zhao-Tian-yi/Learning-to-Pan-sharpening-with-Memories-of-Spatial-Details.git}.
\end{abstract}


\section{Introduction}
With the rapid development of spectral imaging technology, multi-spectral (MS) image plays a vital role in many fields \cite{liu2021gafnet, lu2016joint,wei2023adversarial}, such as climate monitoring, object detection, military reconciliation, and other fields. The significance of remote sensing information is undeniably apparent. For individuals relying on such data, the precision of remote sensing information plays a crucial role in determining the quality of their work and daily life. However, limited by the physical equipment, the onefold sensor cannot guarantee the high spectral and high spatial resolution of the captured images at the same time \cite{vivone2020new,feng2022deep}. Sensors usually obtain either a high-resolution panchromatic (PAN) image or a high-spectral MS image. Therefore, they are usually captured via two different sensors and then fuse a high-resolution multispectral (HRMS) image via pan-sharpening technology \cite{CBD_alparone2007comparison_MRA, zhang2012adjustable,zhou2014gihs}. Thus, the pan-sharpening task can be regarded as a crucial pre-processing step for numerous remote sensing data applications.
\begin{figure}[!tbp]
    \centering
    \includegraphics[width=\columnwidth]{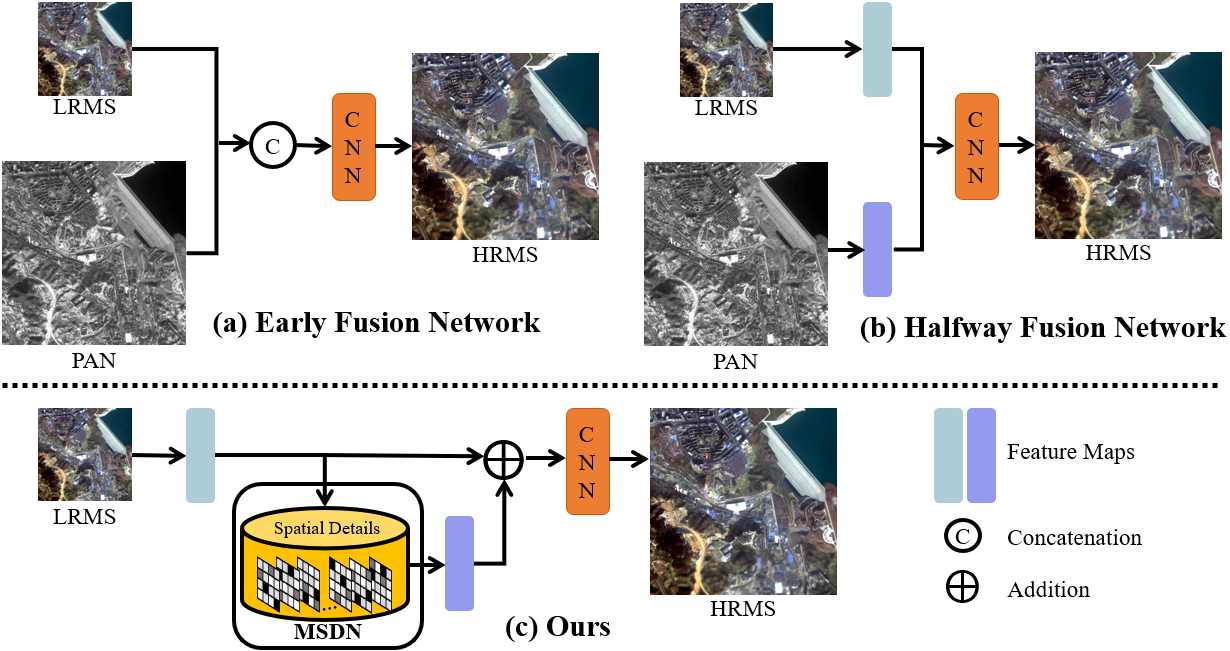}
     % \vspace{-0.2cm}
    \caption{The comparison between existing DL-based methods and our proposed method. In our approach, PAN image is utilized to supervise the learning process of the memory-based module, while is not required during testing.}
    % \vspace{-0.4cm}
    \label{fig1}
\end{figure}

Due to the powerful representation ability of convolutional neural network (CNN), many researchers have utilized CNNs for pan-sharpening \cite{Masi2016PNN,wei2017boosting,zhang2019pan,Cai2021SRPPNN,zheng2020hyperspectral,wang2021frmlnet}. Current deep learning (DL)-based pan-sharpening methods mainly fall into two types: One is the ``early fusion" networks \cite{Masi2016PNN,wei2017boosting,zheng2020hyperspectral,bandara2021hyperspectral} as shown in Figure~\textcolor{red}{\ref{fig1}-(a)}, which adopt a pixel-level fusion way to transfer spatial details from PAN image to MS image. Another is the ``halfway fusion" networks \cite{zhou2022modality,zhou2022effective,Cai2021SRPPNN} as shown in Figure~\textcolor{red}{\ref{fig1}-(b)}, which firstly extract the spatial and spectral details from PAN and MS respectively, and then perform interactive fusion process to obtain HRMS images. It is clear that these two kinds of pan-sharpening methods need the paired PAN and MS image as the input. 
Actually, due to the equipment failure of sensors, the lack of PAN or MS image usually occurs in practise, which will cause these methods to fail. In these scenarios, if we can still perform the pan-sharpening task effectively via only one of the images, it will greatly improve the utilisation of these data and reduce the hardware cost. Several studies \cite{poterek2020deep,liu2022casr} consider coloring PAN images to solve the situation where only PAN images are available. While for MS-only situation, a straightforward way is to apply MS super-resolution \cite{dong2015image,haut2019remote,lei2021hybrid}. However, the image quality cannot be guaranteed compared to the pan-sharpening methods due to the lack of spatial details from PAN. To this end, this article mainly focuses on pan-sharpening using only MS images.
% In other words, these methods will fail when either is missing.
% Actually, it is costly to obtain paired PAN and MS in practical application. In these scenarios, if we can still perform the pan-sharpening task effectively via only the MS image, it will significantly reduce the hardware cost and payload weight.
% A straightforward way is to apply super-resolution for the MS, however, the image quality cannot be guaranteed compared to pan-sharpening methods due to the lack of spatial details from PAN. Therefore, it is necessary to design a new pan-sharpening method that only needs MS image as the input while ensuring sharpening performance. 

\begin{figure}[!tbp]
    \centering
    \includegraphics[width = \linewidth]{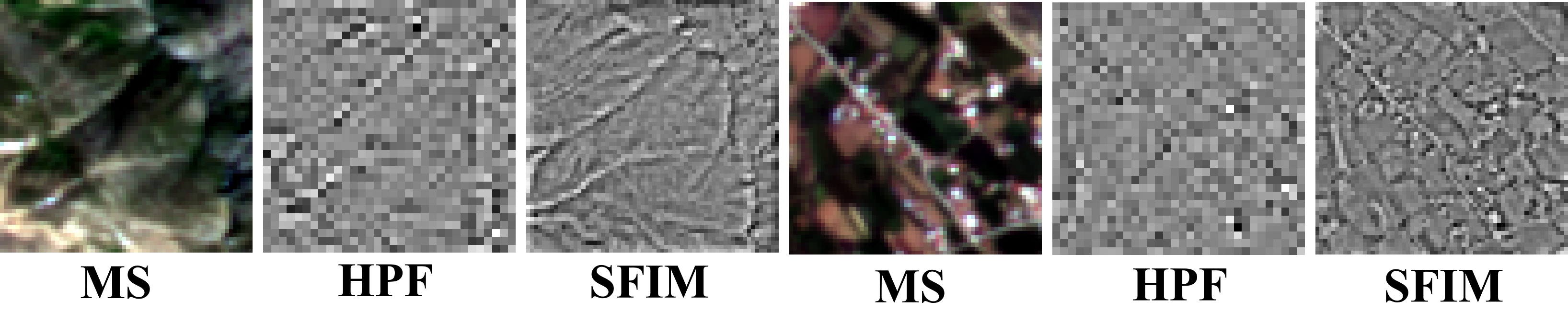}
     \vspace{-0.5cm}
    \caption{The visualization of spatial details $\mathbf P_s$ used in DL-based detail injection models. We visualize $\mathbf P_s$ generated from two commonly hand-designed methods (HPF \cite{chavez1991comparison_HPF} and SFIM \cite{SFIM_liu2000smoothing_MRA}).}
    \label{fig:Ps_visual}
    \vspace{-0.3cm}
\end{figure}

In fact, pan-sharpening is the process of injecting the spatial details from the PAN image to guide the super-resolution of the MS image. We run some classic main-stream pan-sharpening methods and observe that the extracted spatial details from PAN images are mainly high-frequency cues, i.e., the edges reflect the contour of input PAN images. For example, in Figure \ref{fig:Ps_visual}, we visualize the spatial details extracted by HPF \cite{chavez1991comparison_HPF} and SFIM \cite{SFIM_liu2000smoothing_MRA} algorithms.
It can be seen that the spatial details are actually the contour of the PAN image. This motivates us to develop a PAN-agnostic representation to store some base edges, so as to compose the contour of the corresponding PAN image via them. Actually, such a process can be implemented using a memory network \cite{weston2015memory}. During the training phase, the memory-based network can extract and memorize the spatial details and is supervised by PAN images. In the testing phase, we can utilise the memory-based network to construct spatial details rather than directly obtaining them from the corresponding PAN image. In this way, we can perform the pan-sharpening task only with the MS image when inference and the aforementioned problem are tackled.

To this end, a Memory-based Spatial Details Network (MSDN) is designed to replace the PAN image during testing, as shown in Figure~\textcolor{red}{\ref{fig1}-(c)}. To formulate this network, we decompose it and propose two subnetworks: the memory-controlled subnetwork and the weighted coefficient subnetwork. In the memory-controlled subnetwork, we encode the MS features in the memory encoder to query the corresponding spatial details from the memory bank and then obtain the spatial features through the memory decoder. As for the weighted coefficient subnetwork, we utilize channel attention to predict the weighted features. These two subnetworks are combined to memorize the generation of spatial details during training and construct the required spatial details when inference. To validate the effectiveness of the proposed MSDN, we integrate the proposed MSDN module into the existing DL-based pan-sharpening methods and construct an MSDN-based pan-sharpening framework. To better integrate the MSDN into the framework, we redesign an injection network to inject the generated spatial details adaptively. The overall framework is easy to extend and can be trained in an end-to-end manner.
% \vspace{0.2cm}

In summary, our contributions are listed as follows:

\begin{itemize}
    \item We propose a new way to represent spatial details and perform pan-sharpening task using only MS images. To the best of our knowledge, this is the first attempt to generate the spatial details from a memory network instead of the corresponding PAN image.
    \item We design a Memory-based Spatial Details Network consisting of a memory-controlled subnetwork and a weighted coefficient subnetwork to generate the required spatial details from MS features. We also devise a memory-based spatial details learning to memorize the generation of spatial details during training.
    \item We construct an MSDN-based pan-sharpening framework to assess the effectiveness of MSDN. Extensive experiments on the Gaofen1 and WorldView-4 satellites show that our framework achieves state-of-the-art qualitative and quantitative performance.
\end{itemize}

The remainder of the article is organized as follows. In the next section, we review the scientific literature related to our proposed approach. Section \ref{method} presents our motivation and gives a detailed description of the proposed module. Experimental results are presented in Section \ref{experiments}, along with a comparison with state-of-the-art methods, and in Section \ref{conclusion} we conclude with a discussion of our contribution.

\section{Related Works}
% \subsection{Spatial Details Injection Model}
% \noindent\textbf{Traditional Detail Injection Method.} 
\subsection{Traditional Spatial Detail Injection Model}
The general equation for pan-sharpening detail injection model is as follows:
\begin{equation}
    \mathbf{H}=\mathbf{\widetilde{MS}}+g \mathbf{P_s},
    \label{eq:traditional-inject}
\end{equation}
where $\mathbf H$ is the predicted HRMS image after pan-sharpening, $\mathbf{\widetilde{MS}}$ denotes the MS image interpolated at PAN scale, $g$ represents the injection coefficient that controls the injection of the extracted spatial details and $\mathbf P_s$ indicates the extracted spatial details. The traditional detail injection methods are mainly divided into two kinds of methods: CS-based and MRA-based. The difference between these two methods is manifested mainly in the $\mathbf P_s$ representation of Equation~\textcolor{red}{\ref{eq:traditional-inject}}.

In the CS-based detail injection model, $\mathbf P_s = \left(\mathbf P-\mathbf I\right)$ \cite{vivone2014critical,vivone2020new}, where $\mathbf P$ is the PAN image and $\mathbf I$ is the intensity component, generally defined as a weighted sum $\mathbf I=\sum_{i=1}^B \omega_i \mathbf{\widetilde{MS}_i}$. On the basis of this expression, many CS-based pan-sharpening algorithms \cite{garzelli2007optimal, ERGAS_wald2000quality} are proposed by exploring different ways to predict the weights $\omega$ and the injection coefficients $g$.

As for the MRA-based detail injection model, the required spatial details $\mathbf P_s$ can be mainly extracted from PAN image, $\mathbf P_s = \left(\mathbf P-\mathbf P_l\right)$ \cite{vivone2014critical,vivone2020new}, where $\mathbf P_l$ is denoted as the low-pass spatial resolution version of the PAN images. MRA-based methods \cite{GLP_aiazzi2002context_MRA, otazu2005introduction} focus on how to extract the spatial details $\mathbf P_s$ from PAN and estimate the coefficients $g$.
% In brief, the traditional details injection methods focus on the various mathematical representations of $\mathrm P_s$ and $g$.

\begin{figure*}[!t]
    \begin{center}
    \includegraphics[width = 0.82\linewidth]{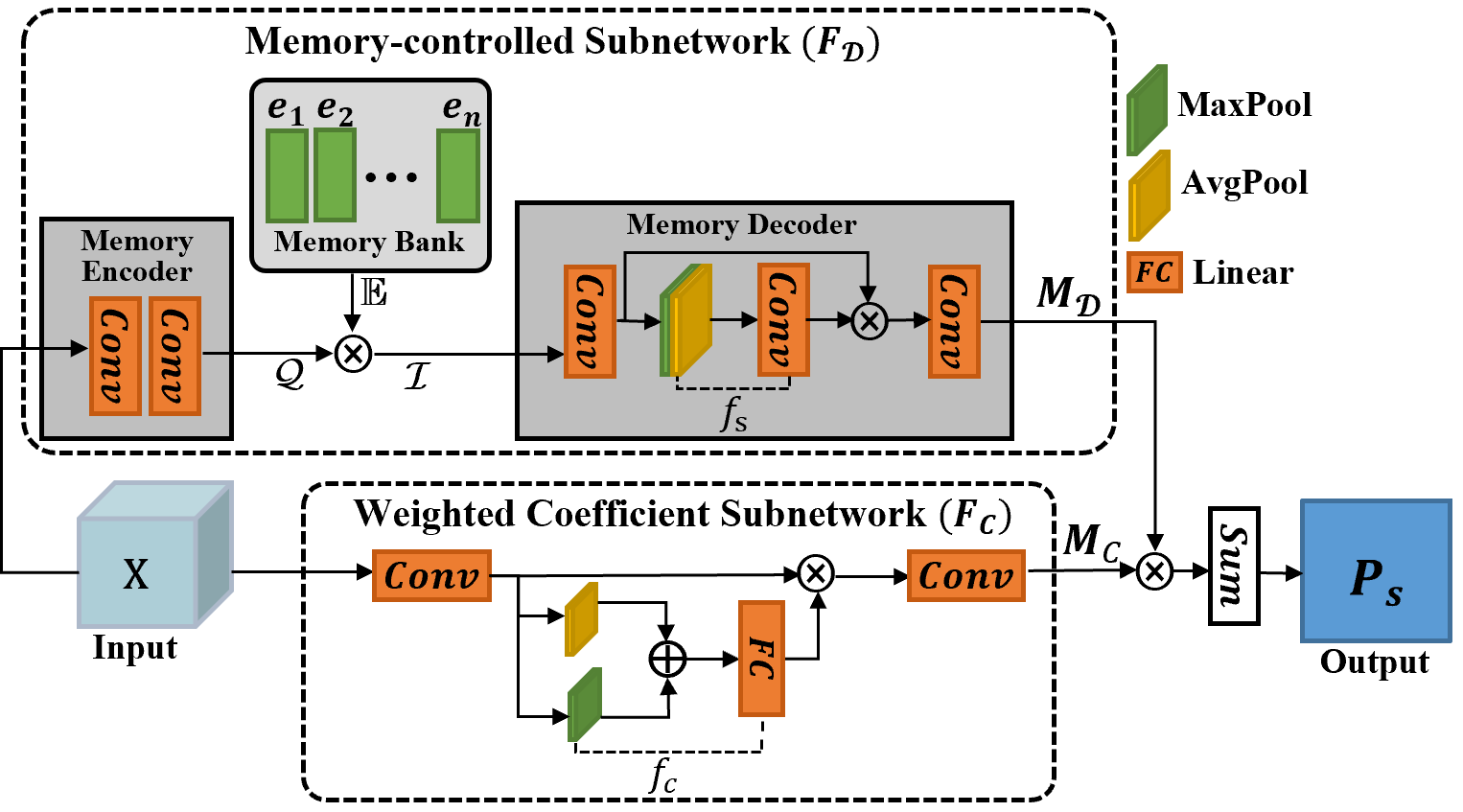}
    \end{center}
     % \vspace{-0.2cm}
    \caption{The overall pipeline of our Memory-based Spatial Details Network. $\bigotimes$ and $\bigoplus$ denote the operation of the element-wise multiplication and addition respectively. For clear illustration, we do not show the activation functions in this module.}
    % \vspace{-0.3cm}
    \label{fig2}
\end{figure*}

% \noindent\textbf{DL-based Detail Injection Model.}
\subsection{DL-based Spatial Detail Injection Model}
With the great success of convolutional networks in computer vision tasks, the DL-based detail injection approaches are also proposed in pan-sharpening based on Equation~\textcolor{red}{\ref{eq:traditional-inject}}. Similar to the traditional methods, the process of some DL-based detail injection methods can be formulated as follows:
% \vspace{-0.1cm}
\begin{equation}
\mathbf H=\mathbf{\widetilde{MS}}+\mathcal{F}_{\theta} \left(\mathbf P_s\right),
\label{eq:dl-inject}
\end{equation}
where $\mathcal{F}_{\theta}$ is the CNNs with the network parameters $\theta$. These methods replace the injection coefficients $g$ by exploring a non-linear mapping function fed by $\mathbf P_s$. For example, DiPAN \cite{he2019pansharpening} is designed from a general detail injection formulation to accommodate conventional MRA-based pan-sharpening methods. Similarly, PANNet \cite{yang2017pannet} introduces the high-pass filter \cite{chavez1991comparison_HPF} into the $\mathcal{F}_{\theta}$ to optimize the spatial details into CNNs. Deng et al. \cite{deng2021detailFusionNet} develop two detail-based architectures according to Equation~\textcolor{red}{\ref{eq:dl-inject}} called CS-Net and MRA-Net to support injection strategy for CNNs. Recently, DIM-FuNet \cite{xiang2022detail} is proposed to bridge classic models and deep neural networks by integrating detail extraction and injection fidelity terms.

The above DL-based detail injection methods focus on designing successive CNNs architecture to predict injection spatial features $\mathcal{F}_{\theta} \left(\mathbf P_s\right)$ while still using the hand-designed methods to obtain the $\mathbf P_s$ from PAN images. However, this way restricts them must use paired PAN and MS images for pan-sharpening. In this paper, we formulate a new $\mathbf P_s$ representation to solve this problem. Thus, we can use only MS images to perform the pan-sharpening task by representing the overall Equation~\textcolor{red}{\ref{eq:dl-inject}} with CNNs in an end-to-end manner.

% Cai et al. \cite{Cai2021SRPPNN} propose a super-resolution guided progressive Pan-sharpening neural network (SRPPNN), which inject multi-scale spatial features to improve the performance.

\subsection{Memory networks} 
Weston et al. \cite{weston2015memory} first propose memory networks for the task of answering questions. From then on, memory-based networks enter the vision of people. Deng et al. \cite{deng2019object} design a long-term memory to memorize the various appearance of objects for video object detection. After that, MGUIT \cite{Xie_2021_CVPR} is proposed to store and propagate instance-level style information for image translation. Similarly, Ji en al. \cite{Ji_2022_CVPR} present a novel memory-based framework that stores blurry to achieve fine-grained video deblurring. Recently, Yan et al. \cite{yan2022memory} propose a novel network for pan-sharpening called MMNet, which combines the benefits of both model-based and DL-based methods to enhance the interpretation of the network. Similar to these methods, in this paper, we also propose a memory-based network called MSDN, which is utilized to replace the traditional hand-designed methods to extract and represent the spatial details $\mathbf P_s$.

\section{Memory-based Spatial Details Network} \label{method}
In this section, we first introduce our motivation in Section \ref{motivation}. According to our motivation, we propose the Memory-controlled Subnetwork (Section  \ref{memorycontrolled}) and Weighted Coefficient Subnetwork (Section  \ref{weightedcoefficient}). Finally, we present a Memory-based Spatial Details Learning in Section \ref{spatial-details-learning}.

\subsection{Motivation} \label{motivation}
As discussed above, the pan-sharpening task can be considered as a process of injecting spatial details of PAN into MS super-resolution. Existing DL-based detail injection models usually rely on traditional hand-designed methods to obtain spatial details. From Figure~\textcolor{red}{\ref{fig:Ps_visual}}, we can observe that these spatial details obtained from hand-designed methods are mainly high-frequency cues, such as prominent outlines and points. This motivates us whether we can obtain these spatial details generated from the sample-invariant model instead of obtaining them from the corresponding PAN image in DL-based methods. To achieve it, referring to the sparse coding theory \cite{zhu2012sparse,maeda2022image}, we consider utilizing the form of a weighted combination dictionary model to decompose the representation of spatial details $\mathbf P_s$:
\begin{equation}
\mathbf P_s=\sum_{n=1}^N \mathcal D_n \mathcal C_n=\mathcal D \mathcal C,
\label{equ:Ps}
\end{equation}
where $\mathcal D$ is the spatial dictionary, which needs to be optimized from a large number of spatial detail image patches during training and finally utilized to represent the features of these patches when inference. As for $\mathcal C$, it is used to sparsely represent the coefficient of these features. In this way, we can construct the $\mathbf P_s$ for a given PAN image. 

% and $\mathcal C$ is the weighted coefficient. Spatial dictionary $\mathcal D$ needs to be optimized from a large amount of spatial details patches so that it can sparsely represent the features.
% Therefore, during the training process, we can utilize PAN to supervise the memory network to learn how to generate $\mathbf P_s$, while reconstructing $\mathbf{P_s}$ through the network without PAN images during the inference stage.  

\subsection{Memory-controlled Subnetwork ($\mathcal{F}_\mathcal D$)} \label{memorycontrolled}
To model the spatial dictionary $\mathcal D$, we consider it is not a simple set of dictionaries but a memory network containing $N$ slot learnable spatial items. The principle of this memory network $\mathcal D$ is to learn a controller that can dynamically access the relevant spatial features given the upsampled MS features. As shown in Figure~\textcolor{red}{\ref{fig2}}, the whole subnetwork consists of memory bank, memory encoder, and memory decoder, which are introduced as follows.
% To facilitate the subnetwork $\mathcal D$, a memory encoder and a memory decoder are also established to query and readout the corresponding detail features, which are introduced as follows. 

\noindent\textbf{Memory bank.} 
The memory bank stores the spatial items $\mathrm{E}=\{e_1,e_2...e_n\} \in \mathbb{R}^{N \times s^2}$ in the latent embedding space, and each of them is a learnable embedding, where $s$ is an up-scaling factor. To reduce the size of the memory bank, we fix the dimension of $\mathcal{E}$ and use the repeat operation to expand it to the required dimension:
\begin{equation}
    \mathbb{E} \in \mathbb{R}^{N \times H \times W} = \Gamma_{1\times H \times W}\left(\mathrm{E} \right),
\end{equation}
where $\Gamma_{a \times b \times c}(\cdot)$ denotes the $a \times b \times c$ repeat operations. The weight parameters of the spatial items are initialized with Kaiming initialisation \cite{he2015delving} and are updated through the overall training loss of the network.

\noindent\textbf{Memory encoder.} 
The memory encoder is utilized to convert the input features into the form of 
the query features required by the memory bank. Specifically, assuming that the input upsampled MS features are $\mathrm{X} \in \mathbb{R}^{C \times H \times W}$, we employ two consecutive convolutional layers to generate the query features. The generation process is defined as:
\begin{equation}
    \mathcal Q = \mathcal{W}_q * \mathrm{X},
    \label{equ:query-encoder}
\end{equation}
where $\mathcal{W}_q$ denotes the two consecutive convolutions with kernel sizes $3\times3$ and $1\times1$ respectively. The query feature maps are denoted as $\mathcal Q \in \mathbb{R}^{N \times H \times W}$.

\noindent\textbf{Memory decoder.}
After obtaining the query features $\mathcal Q$, we multiply it with the spatial items $\mathbb{E}$ as input and utilize the memory decoder to extract the spatial features. The input of the memory decoder can be expressed as:
\begin{equation}
    \mathcal I \in \mathbb{R}^{N \times H \times W} =\mathbb{E} \otimes \mathcal Q,
    \label{equ:input-decoder}
\end{equation}
where $\otimes$ is the element-wise multiplication. In the memory decoder, we first acquire a new feature representation by utilizing a $3\times3$ convolution $\mathcal{W}_c$. To enrich the representation of spatial features, a spatial attention operation $f_s$ is deployed to predict the feature weights. Finally, we multiply weights and pass through a $1\times1$ convolution $\mathcal{W}_d$ to obtain the spatial features $\mathcal{M}_{\mathcal D} \in \mathbb{R}^{C \times H \times W}$. Formally, the process can be defined as:
\begin{equation}
    \mathcal{M}_{\mathcal D} = \mathcal{F}_\mathcal D(\mathcal I)= \mathcal{W}_d * \left(\left(\mathcal{W}_c*\mathcal I\right) \otimes f_{s}\left(\mathcal{W}_c*\mathcal I\right)\right).
    \label{equ:memory-decoder}
\end{equation}
% where $\otimes$ is the element-wise multiplication and $\mathcal{M}_{\mathcal D} \in \mathbb{R}^{C \times H \times W}$.

\subsection{Weighted Coefficient Subnetwork ($\mathcal{F}_\mathcal C$)} \label{weightedcoefficient}
As discussed above, the dictionary $\mathcal D$ is modeled as a memory-based network. To
obtain the representation of spatial details $\mathbf P_s$, we also need to model the sparse coefficient $\mathcal C$. Thus, according to the definition of sparse coding in super-resolution \cite{yang2010image}, we formulate the MS super-resolution as follows:
% \vspace{-0.4mm}
\begin{equation}
    \begin{aligned}
        & \mathbf{\widetilde{M S}}=\mathcal{D}_l \alpha, \\
        & \mathbf H =\mathcal{D}_h \alpha,
        \label{equ:ms-superresolution}
    \end{aligned}
\end{equation}
where $\alpha$ is the common sparse coefficients, $\mathcal{D}_l$ and $\mathcal{D}_h$ are spatial dictionaries extracted from $\mathbf{\widetilde{M S}}$ and $\mathbf{H}$, respectively. 
According to Equation~\textcolor{red}{\ref{equ:ms-superresolution}}, $\mathbf{\widetilde{M S}}$ and $\mathbf H$ can deduce each other, as follows:
\begin{equation}
    \begin{aligned}
         \mathbf H = {\mathcal{D}_h}{\mathcal{D}_l}^{-1}\mathbf{\widetilde{M S}},
        \label{equ:ms-superresolution-elimination}
    \end{aligned}
\end{equation}
substituting  Equation~\textcolor{red}{\ref{equ:Ps}} and Equation~\textcolor{red}{\ref{equ:ms-superresolution-elimination}} into Equation~\textcolor{red}{\ref{eq:traditional-inject}}, we obtain:
\begin{equation}
    \begin{aligned}
    {\mathcal{D}_h}{\mathcal{D}_l}^{-1}\mathbf{\widetilde{M S}} &=\mathbf{\widetilde{M S}} + g \mathcal D \mathcal C, \\
    g \mathcal D \mathcal C  &= ( {\mathcal{D}_h}{\mathcal{D}_l}^{-1}-1)\mathbf{\widetilde{M S}},  \\
    \mathcal C &= \mathcal{D}^{-1}g^{-1}\left({\mathcal{D}_h}{\mathcal{D}_l}^{-1}-1\right) \mathbf{\widetilde{M S}}.
    \label{eq:gamma_representation}
    \end{aligned}
\end{equation}

Since $\mathcal{D}^{-1}g^{-1}$ are the matrix operation and $\left({\mathcal{D}_h}{\mathcal{D}_l}^{-1}-1\right) \mathbf{\widetilde{M S}}$ can be viewed as the operation of extracting the $\mathbf{\widetilde{M S}}$ features, the complex representation $\mathcal{D}^{-1}g^{-1}\left({\mathcal{D}_h}{\mathcal{D}_l}^{-1}-1\right)$ can be solved with CNNs. Specifically, considering the sparse characteristics of $\mathcal C$, we propose a weighted coefficient subnetwork ($\mathcal{F}_\mathcal C$) to predict it. In this subnetwork, a channel attention module $f_c$ is utilized to predict the weighted features, and two $1\times1$ convolutions ($\mathcal{W}_a$ and $\mathcal{W}_k$) are used to transform the dimensions of the features. Thus, for the input upsampled MS features $x$, the output features $\mathcal{M}_{\mathcal C} \in \mathbb{R}^{C \times H \times W}$ can be regarded as the coefficient $\mathcal C$, which can be represented as: 
\begin{equation}
\mathcal{M}_{\mathcal C}  = \mathcal{F}_\mathcal C (x) = \mathcal{W}_k * (f_c(\mathcal{W}_a * \mathrm{X})\otimes (\mathcal{W}_a * \mathrm{X})).
\label{eq:gamma_cnnrepresentation}
\end{equation}

\subsection{Memory-based Spatial Details Learning} \label{spatial-details-learning}
With the outputs of $\mathcal{M}_{\mathcal D}$ and $\mathcal{M}_{\mathcal C}$, we perform the element-wise multiplication and then sum over the first dimension to obtain the $\mathbf P_s$, which can be formulated as:
\begin{equation}
\mathbf P_s \in \mathbb{R}^{1 \times H \times W} = Sum\left(\mathcal{M}_{\mathcal D} \otimes \mathcal{M}_{\mathcal C},dim=1 \right).
\label{eq:final-ps}
\end{equation}
To maintain the sparsity of the output features $\mathcal{M}_{\mathcal C}$, L1 regularization is used during training to limit the optimization process of $\mathcal{M}_{\mathcal C}$, which defines as:
\begin{equation}
    \mathcal{L}_{norm} = \sum_{i=1}^H \sum_{j=1}^W \sum_{k=1}^C \Vert\mathcal{M}_{\mathcal C_{ijk}} \Vert_1.
\label{eq:regulazation}
\end{equation}
For the MSDN to memorize the $\mathbf P_s$ generation process through the memory-controlled subnetwork $\mathcal D$ during training, we utilize the high-pass details $\mathbf{HP}$ \cite{yang2017pannet} computed by subtraction between PAN and low-resolution PAN (obtained by using an averaging filter) as the ground truth to supervise the spatial details learning. To guide the generated $\mathbf P_s$ to be similar to $\mathbf{HP}$ in their distribution space, we exploit the Kullback-Leibler divergence $D_{KL}\left(\cdot \right)$ to measure and minimise the difference between two distributions. Thus, a spatial details memorizing loss $\mathcal L_{Mem}$ is designed and can be defined as follows:
\begin{equation}
\mathcal L_{Mem} = D_{KL}\left( \mathbf{HP} || \mathbf P_s\right) + \mathcal{L}_{norm}.
\end{equation}
To facilitate the calculation, we stretch two spatial detail features~($\mathbf{HP}$ and $ \mathbf P_s$)~as vectors. Through the $\mathcal L_{Mem}$ loss, the values of the spatial items are almost constant during training and the generation of spatial details is gradually memorized by the MSDN.
% With the spatial details memorizing loss, we find that the contrast of the output spatial items becomes stronger as the training progresses, and the items are almost constant during the last three quarters of training. The visualization of constant spatial items is shown in Figure~\textcolor{red}{\ref{fig:dictionary_visualization}}.

\section{MSDN-based Pan-sharpening Framework}
To evaluate our MSDN module, we construct a pan-sharpening framework incorporating MSDN. In this framework, we replace the traditional generation process of $\mathbf P_s$ with MSDN and feed the $\mathbf P_s$ into the network $\mathcal{F}_{\theta}$ (in Equation~\textcolor{red}{\ref{eq:dl-inject}}) to inject spatial features. The detailed descriptions of the framework structure and the loss functions are as follows.

% we also re-design the injection network $\mathcal{F}_{\theta}$ and use it to inject generated $\mathbf P_s$. The detailed structure of framework and the description of the loss functions are as follows.

\begin{figure*}[!t]
    \begin{center}
    \includegraphics[width=0.7\linewidth]{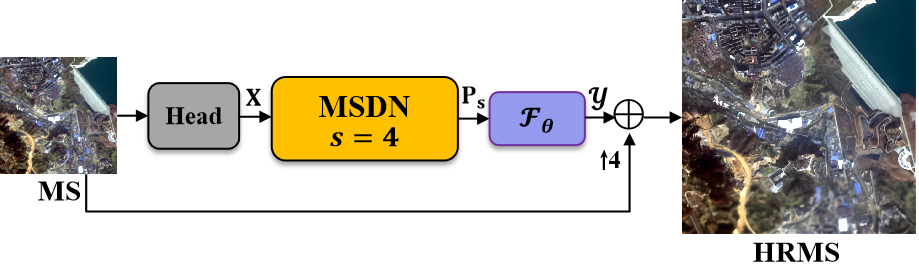}
    \end{center}
    % \vspace{-0.5cm}
        \vspace{-0.3cm}
    \caption{The architecture of MSDN-based pan-sharpening framework.~$\uparrow$ represents bicubic upsampling operation.}
    \label{fig3}

\end{figure*}

\subsection{Overall Framework} \label{framework}
As shown in Figure~\textcolor{red}{\ref{fig3}}, our MSDN works as a plug-in mode and is injected into an existing pan-sharpening framework \cite{deng2021detailFusionNet}. In this framework, for the input MS image, stacked residual blocks denoted as ${Head}$ are used to extract MS features. We utilize MSDN after ${Head}$ to replace the original spatial details $\mathbf P_s$ generation:
\begin{equation}
    \mathbf P_s = \mathcal{F}_{\mathrm{MSDN}}\left(\mathrm{Head} \left(\mathbf{MS}\right)\right),
\end{equation}
where $\mathcal{F}_{MSDN}$ denotes the MSDN. The generated spatial details $\mathbf P_s$ is then fed into the network $\mathcal{F}_{\theta}$ to obtain the injection features $\mathcal Y = \mathcal{F}_{\theta} \left(\mathbf P_s\right)$. 
\begin{align}
    \mathbf{H} &= \mathbf{\widetilde{M S}} + \mathcal Y  \\
                &= \mathbf{\widetilde{MS}}+\mathcal{F}_{\theta} \left(\mathbf P_s\right) \nonumber,
\end{align}
In this framework, we set $s=4$ in MSDN to obtain HRMS since the ratio of the spatial resolution between the PAN and the MS is 4. Note that other sizes of HRMS ($\times2$ and $\times8$, etc.) can also be generated with our MSDN.

% \textcolor{red}{It seems Equation(14) is not consistent with Equation(2), the extra term with Equation(2) is R2(R1(MS). is this term related with the Ref[22]), please explain this point.}
% According to the Equation(\textcolor{red}{\ref{eq:dl-inject}}), in each stage, we also perform the addition operation between the output and the upsampling MS. Overall, the MSDMNet can be summarized as follows:
% \begin{equation}
% \left[\mathrm{H}_{2 \times}, \mathrm{H}_{4 \times}\right]=\mathrm{MSDMNet}(\mathrm{MS}),
% \end{equation}
% where $\mathrm{H}_{2 \times}$ and $\mathrm{H}_{4 \times}$ denote the output of MSDMNet with different scales.

\subsection{Re-designed Injection Network}
\begin{figure}[!t]
    \centering
    \includegraphics[width=\linewidth]{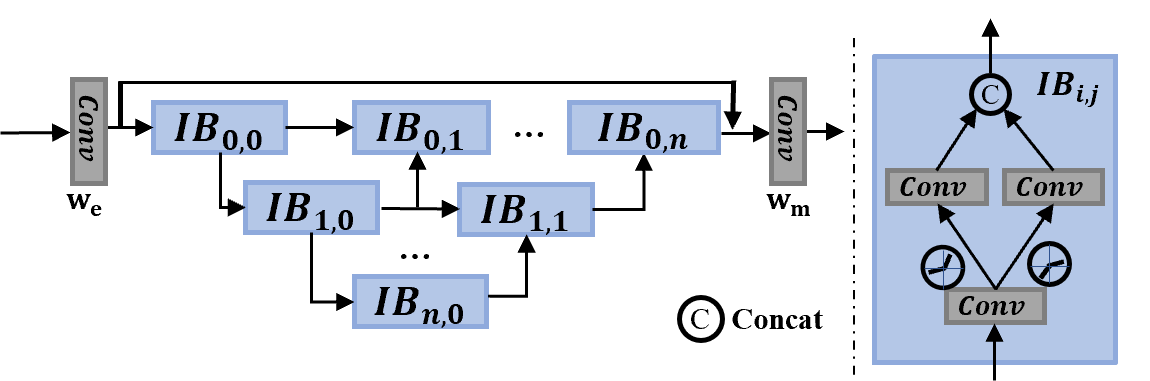}
    \caption{The flowchart of Nested Injection Network.}
    \label{fig4}
    % \vspace{-0.3cm}
\end{figure}

For injection network $\mathcal{F}_{\theta}$, we can use the existing methods to inject the generated $\mathbf{P_s}$. To better adapt the MSDN to the framework, we redesign the $\mathcal{F}_{\theta}$ and propose the Nested Injection Network (NIN). The structure of the NIN is shown in Figure~\textcolor{red}{\ref{fig4}}. Since the U-shaped structure \cite{ronneberger2015u} performs excellently on the pixel-wise task, we utilize this structure with Injection Block~(IB) as a basic block. In IB, each block contains two \emph{PReLU} \cite{he2015delving} and three convolutions to improve the feature representation by exploring both positive and negative features. In specific, for the input $\mathbf P_s$, we first acquire a new feature map $y_d \in \mathbb{R}^{C \times H \times W}$ by utilizing a convolution layer $\mathcal{W}_e$, $y_d=\mathcal{W}_e* \mathbf P_s$. To obtain the representation of both positive and negative features, we deploy \emph{PReLU} and convolution layers on $y_d$ and $-y_d$ in each block:

    \begin{align}
        & {x}_p \in \mathbb{R}^{C/2 \times H \times W} = \mathcal{W}_p*\sigma(y_d),\\
        & {x}_n \in \mathbb{R}^{C/2 \times H \times W} = \mathcal{W}_n*\sigma(-y_d).
    \end{align}

where $\sigma$ denotes the \emph{PReLU} activation, $\mathcal{W}_p$ and $\mathcal{W}_n$ are the $3\times3$ convolution layers. We concat ${x}_p$ and ${x}_n$ together as each output of IB.
% We then concat them to acquire the each output of ICB: 
% \begin{equation}
% y \in \mathbb{R}^{C \times H \times W} =Cat\left( \mathcal{F}_p , \mathcal{F}_n \right).
% \end{equation}
The final output of NIN can be formulated as:
\begin{equation}
\mathcal Y \in \mathbb{R}^{4 \times H \times W} =y_d+\mathcal{W}_m*f_{IB}(y_d),
\end{equation}
where $f_{IB}$ represents the U-shaped IB group and $\mathcal{W}_m$ is the final $1\times1$ convolution layer.

\subsection{Loss Functions} 
As mentioned above, the overall loss function consists of two parts: one is the mean absolute error loss ($\mathcal L_{1}$) aims at reconstructing the HRMS image with the ground-truth MS image and the other is the spatial details memorizing loss ($\mathcal L_{Mem}$) which is used to supervise MSDN to reconstruct spatial details $\mathbf P_s$. The final loss function of our framework is written as follows:
\begin{equation}
\begin{aligned}
    &\mathcal L_{1} = \left\|\mathrm{H}-\mathrm{GT}\right\|_1, \\
    &\mathcal L =\mathcal L_{1} +\lambda \mathcal L_{Mem},
\end{aligned}
\end{equation}
where $\mathrm{H}$ is the predicted HRMS, GT is the ground-truth HRMS and $\lambda$ is a hyper-parameter for balancing ($\mathcal L_{1}$) and ($\mathcal L_{Mem}$) loss functions. In our setting, ~$\lambda$~is set as 0.001, such that the average gradient of each loss is at the same scale. 

\begin{table*}[]\scriptsize
\centering
\caption{Description of the pan-sharpening quality metrics.}
    \label{tab:Metrics}
    \renewcommand{\arraystretch}{3.8}
\begin{tabular}{c:c:l:c}
\hline

 \textbf{Metrics} & \textbf{Full name}     & \multicolumn{1}{c:}{\textbf{Formula}}     &\textbf{Nomenclature}                        \\ \hline

 \multirow{2}{*}{\textbf{ERGAS}}                & \multirow{2}{3cm}{Erreur Relative Global Adimensionnelle de Synthese} & \multirow{1}{*}{$    ERGAS=100\cdot\frac{h}{l}\sqrt{\frac{1}{K}\sum\limits_{k=1}^{K}\left(\frac{RMSE\left(X_k,Y_k\right)}{E\left[X_k\right]}\right)^2}$ }  & \multirow{5}{6.1cm}{$X$: : the predicted fused images; \par 
$Y$: the reference image;\par 
$X^{\prime}$: the weighted sum of each band of MS image; \par 
$Y^{\prime}$: the weighted sum of each band of PAN image;\par 
$K$: the number of image channels;\par 
$E[\cdot]$: the average value of all pixel;\par 
$\mathcal{L}$: the Laplace operator; \par 
$h$: spatial resolutions of PAN image; \par 
$l$: the spatial resolutions of MS image;\par 
$\sigma$:the standard deviation; \par 
$\mu$:the mean value;\par 
$\left|\left| \cdot \right|\right|$: $L_2$ normalization;  \par
$<\cdot,\cdot>$: the inner product. } \\
  & &\multirow{1}{*}{ $RMSE\left(X,Y\right)=\sqrt{E\left[\left(X-Y\right)^2\right]}$ } \\\cline{1-3} 
  
 \textbf{SAM}  & Spectral Angle Mapper  & \multirow{1}{*}{$SAM(X,Y)=E\left[ arccos\left(\sum\limits_{k=1}^K \frac{<X_k,Y_k>}{\left|\left|X_k\right|\right|_2\cdot\left|\left|Y_k\right|\right|_2} \right)\right]$ }                  \\\cline{1-3}   
 
  \textbf{SCC}   & Spatial Correlation Coefficient   &\multirow{1}{*}{    $    \mathrm{SCC} = \frac{E\left[\left(\mathcal{L}\ast X-\mu_{X^\prime}\right)\left(\mathcal{L}\ast Y-\mu_{Y^\prime}\right)\right]} {\sqrt{E\left[\left(\mathcal{L}\ast X-\mu_{X^\prime}\right)^2\right] E\left[\left(\mathcal{L}\ast Y-\mu_{Y^\prime}\right)^2\right]}} $}         \\\cline{1-3} 
   
\multirow{1}{*}{\textbf{Q4}}  &\multirow{1}{3cm}{Unique Score index Q4 for 4-band MS images}   & \multirow{1}{*}{$Q4=Q\left(X^{\prime}, Y^{\prime}\right)=\frac{\sigma_{X^{\prime}Y^{\prime}}}{\sigma_{X^{\prime}}\ \sigma_{Y^{\prime}}}\cdot\frac{2\ \mu_{X^{\prime}}\mu_{Y^{\prime}}}{\mu_{X^{\prime}}^2+\mu_{Y^{\prime}}^2}\cdot\frac{2 \sigma_{X^{\prime}}\sigma_{Y^{\prime}}}{\sigma_{Y^{\prime}}^2+\sigma_{Y^{\prime}}^2}$}   \\ \cline{1-4} 
 
\textbf{QNR}                & Quality with No-Reference        & \multirow{1}{*}{ $    {QNR} =\left(1-D_\lambda\right)^\alpha\left(1-D_S\right)^\beta$  }      & \multirow{3}{6cm}{$\mathbf{P}$: the original PAN images; \par 
$\mathbf{P}_l$: the low-pass spatial resolution of PAN images; \par 
$p$: the coefficient to enhance the spectral differences;\par 
$q$: the coefficient to enhance the spatial differences; \par 
$d_{i, j}(\mathbf{M S},  \mathbf H)= Q\left(\mathbf{M S}_i, \mathbf{M S}_j\right)-Q\left(\mathbf H_i, \mathbf H_j\right)$;\par 
$\alpha$ and $\beta$: hyperparameter, usually set to 1.  }         \\\cline{1-3} 
 
 $\mathbf{D_\lambda}$               & Spectral Distortion Index    &  \multirow{1}{*}{  $    {D_\lambda}=\sqrt[p]{\frac{1}{K\left(K-1\right)}\sum\limits_{i=1}^{K}\sum\limits_{j=1,j\neq i}^{K}\left|d_{i, j}(\mathbf{M S}, X)\right|^p}$        }            \\\cline{1-3} 
 
 $\mathbf{D_s}$               & Spatial Distortion Index       &\multirow{1}{*}{$ {D_S} =\sqrt[q]{\frac{1}{K}\sum\limits_{i=1}^{K}\left|Q\left(X_i,\mathbf{P}\right)-Q\left(\mathbf{M S}_i,\mathbf{P}_l \right)\right|^q}$  }     \\\hline          
\end{tabular}
\end{table*}

\begin{table}[!t]\normalsize
    \begin{center}
        \caption{Ablation experiments for components in our framework on the Gaofen1 dataset. The best results are highlighted in \textbf{bold}.}
    \label{tab:Ablation-1}
    \renewcommand{\arraystretch}{1.3}
    \setlength{\tabcolsep}{2mm}
    \begin{tabular}{cc|cccc}
        \hline
        \textbf{MSDN} & \textbf{NIN} & \textbf{SAM}$\downarrow$& \textbf{ERGAS}$\downarrow$& \textbf{SCC}$\uparrow$ & \textbf{Q4}$\uparrow$ \\  \hline
           &   & 0.117  & 7.530   & 0.886    & 0.658    \\
        \checkmark  &   & 0.102   & 6.607    & 0.906 & 0.703       \\
        \checkmark  & \checkmark   & \textbf{0.099}   & \textbf{6.320}   & \textbf{0.912}     & \textbf{0.705}   \\
        \hline
    \end{tabular}
    \end{center}
    % \vspace{-0.3cm}
\end{table}
    
\section{Experiments} \label{experiments}
In this section, we first describe the datasets and evaluation metrics used in our experiments. Then, an exhaustive investigation of ablation studies is presented. Next, we visualize the intermediate results of our method. Finally, to illustrate the effectiveness of our method, comparison experiments were conducted with some closely related methods at reduced resolution data and full resolution data, respectively. Moreover, since our method only requires MS image during in the inference like the MS image super-resolution method, we additionally conduct comparison experiments with the promising super-resolution method to illustrate the effectiveness of the method from the side.

\begin{table}[!t]\normalsize
    \begin{center}
    \caption{Ablation experiment for the sizes of the memory bank on the Gaofen1 dataset. The best results are highlighted in \textbf{bold}.}
        \label{tab:Ablation-4}
    \renewcommand{\arraystretch}{1.3}
    \setlength{\tabcolsep}{2mm}
    \begin{tabular}{c|cccc}
        \hline
        
       \textbf{Size $\mathbf N$} & \textbf{SAM}$\downarrow$ & \textbf{ERGAS}$\downarrow$ & \textbf{SCC}$\uparrow$ & \textbf{Q4}$\uparrow$ \\  \hline
        \textbf{16}  & 0.101 &	6.331 &	0.910 &	0.703    \\
        \textbf{32}  & 0.100 &	6.400 &	0.911 &	0.703  \\
        \textbf{64}  & \textbf{0.099} &	\textbf{6.320} &	\textbf{0.912} &	\textbf{0.705}   \\
        \textbf{128} & 0.100 &	6.422 &	0.908 &	0.699 \\
        \hline
    \end{tabular}
    \end{center}
        \vspace{-0.3cm}
\end{table}

\subsection{Datasets and Metrics}
\noindent\textbf{Datasets.} In this work, all the experiments are conducted on the released pan-sharpening datasets \cite{datasets_meng2020large}, where each satellite includes urban, green vegetation, water scenario, and unlabelled with mixed feature data. We choose Gaofen-1 and WorldView-4 satellites for the experiments. For Gaofen1, the spatial resolution of the PAN images and MS images is 2m and 8m, respectively. For WorldView-4, the spatial resolutions of PAN images and MS images are 0.31m and 1.24m, respectively. In both datasets, the initial sizes of the PAN and MS images are $1024 \times 1024 \times 1$ and $256 \times 256 \times 4$. In order to obtain more training samples, we crop them into image patches of size $256 \times 256 \times 1$ and $64 \times 64 \times 4$ and take these as original images. The MS bands are composed of four standard colors (RGB and near-infrared). Following the Wald’s protocol \cite{wald'sprotocol1997fusion}, the PAN images and the MS images are downsampled by a low-pass filter into image patches of size $64 \times 64 \times 1$ and $16 \times 16 \times 4$, respectively. Furthermore, they are divided into the training set with 90\% pairs and the test set with 10\% pairs, and the original MS images with a size of $64 \times 64 \times 4$ are used as the ground truth. 

\noindent\textbf{Metrics.} The performance assessment is conducted both at reduced and full resolution. For the reduced-resolution, five commonly used evaluation metrics are adopted for evaluating the fusion performance, including SCC \cite{SCC_zhou1998wavelet}, ERGAS \cite{ERGAS_wald2000quality}, SAM \cite{SAM_yuhas1992discrimination} and Q4 \cite{Q4_alparone2004global}. For the full-resolution, since there are no GT images to assess the performance, we apply QNR, $\mathrm D_{\lambda}$ and $\mathrm D_s$ indexes \cite{QNR_alparone2008multispectral} for the quality evaluation. The detailed information is presented in Table \ref{tab:Metrics}.

\begin{table}[!t]\normalsize
    \begin{center}
    \caption{Ablation experiment for the depth of NIN layers in our framework. The best results are highlighted in \textbf{bold}.}
    \label{tab:Ablation-3}
    \renewcommand{\arraystretch}{1.3}
    \setlength{\tabcolsep}{2mm}
    \begin{tabular}{c|cccc}
        \hline
        
        \textbf{Depth of NIN}  & \textbf{SAM}$\downarrow$& \textbf{ERGAS}$\downarrow$& \textbf{SCC}$\uparrow$ & \textbf{Q4}$\uparrow$ \\  \hline
        \textbf{1}   & 0.101  &  6.371  &   0.908  & 0.702     \\
        \textbf{2}   & 0.100   & 6.333   &  0.911   & 0.703 \\
        \textbf{3}   & \textbf{0.099}  &  \textbf{6.320}  &   \textbf{0.912}  & \textbf{0.705}   \\
        \textbf{4}   & \textbf{0.099}  & 6.365   & 0.910    & 0.699    \\

        \hline
    \end{tabular}
    \end{center}
\end{table}

\begin{table}[!t]\normalsize
    \begin{center}
    \caption{Quantitative results to verify the versatility of MSDN.}
    \label{tab:Ablation-2}
    \renewcommand{\arraystretch}{1.3}
    \setlength{\tabcolsep}{2mm}
    \begin{tabular}{cc|cccc}
        \hline
        $\mathbf{\times2}$ & $\mathbf{\times4}$ & \textbf{SAM}$\downarrow$   & \textbf{ERGAS}$\downarrow$ & \textbf{SCC}$\uparrow$   & \textbf{Q4}$\uparrow$    \\ \hline
          & \checkmark & 0.099 &	6.320 &	0.912 &	0.705  \\
        \checkmark & \checkmark & \textbf{0.098} & \textbf{6.293} & \textbf{0.913} & \textbf{0.706} \\ \hline
    \end{tabular}
    \end{center}
\end{table}

\begin{figure}[!t]
    \centering
    \includegraphics[width=0.98\linewidth]{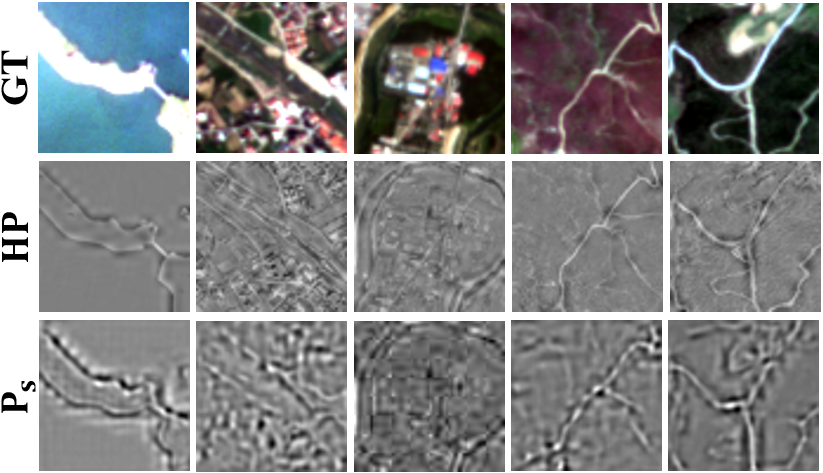}
    \caption{Visualization of generated spatial details $\mathbf{P_s}$. HP is the high-pass images extracted from PAN images.}
    \label{fig:DRN effect}
\end{figure}

\begin{figure}[!t]
    \centering
    \includegraphics[width=1.0\linewidth]{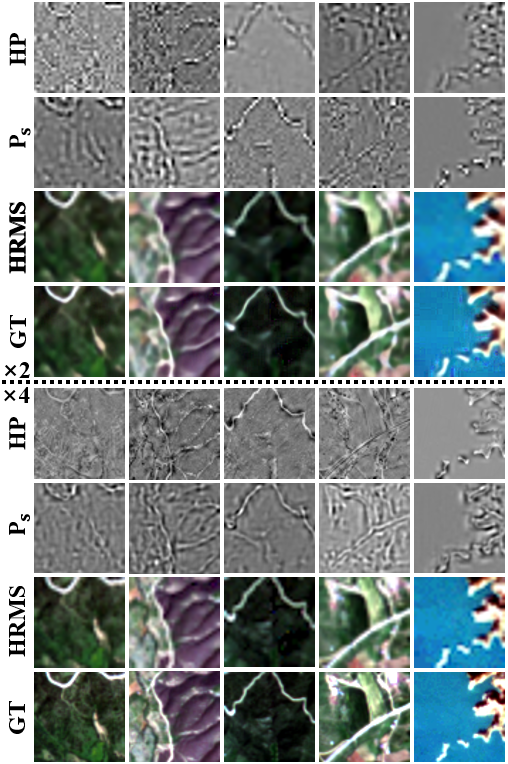}
    % \vspace{0.2cm}
    \caption{Multiscale Outputs visualization. $\times 2$ indicate double up-sampling ($16\times16 \rightarrow 32\times 32$) and $\times 4$ indicate quadruple up-sampling($16\times16 \rightarrow 64\times 64$).}
    \label{fig:Multiscale visual}
    % \vspace{-0.3cm}
\end{figure}

\subsection{Implementation details and Benchmarks}
 \noindent\textbf{Implementation details.} 
For a fair comparison, all experiments are conducted on a single Nvidia GTX 2080Ti GPU and the same Pytorch \cite{paszke2019pytorch} environments with 12GB memory. During training, our methods optimized by Adam \cite{kingma2014adam} optimizer over 200 epochs with a batch size of 16. Horizontal and vertical flipping are also adopted for data augmentation. The learning rate is initialized with $4 \times 10^{-4}$ and decayed by multiplying 0.5 when reaching 50 epochs. For other comparison methods, we apply the default settings in related papers and codes.
During inferencing, our method accomplishes the pan-sharpening task without PAN images. 
 
\noindent\textbf{Benchmarks.} We compare our method with several representative state-of-the-art methods. For pan-sharpening: Four CS-based methods, GS \cite{GS_laben2000process_CS}, GSA \cite{GSA_aiazzi2007improving_CS}, Brovey \cite{Brovey_gillespie1987color_CS} and GFPCA \cite{8075405_GFPCA}; Three MRA-based methods: SFIM \cite{SFIM_liu2000smoothing_MRA}, MTF-GLP \cite{GLP_aiazzi2002context_MRA} and MTF-GLP-HPM \cite{MTF-GLP-HPM-lee2009fast}; Five DL-based methods: PNN \cite{Masi2016PNN}, PANNet \cite{yang2017pannet}, MSDCNN \cite{Yuan2018AMSDCNN}, FusionNet \cite{deng2021detailFusionNet} and ADKNet \cite{Peng2022ADKNet}. As for remote sensing Super-Resolution: Five DL-based methods, SRCNN\cite{dong2015image}, VDSR\cite{kim2016accurate}, LGCNet\cite{lei2017super}, DCM\cite{haut2019remote} and HSENet\cite{lei2021hybrid}. 

\subsection{Ablation experiments}
 \noindent\textbf{Ablation on our framework.} We ablate the components in our constructed framework to investigate the effectiveness of these designs. The baseline is our framework without MSDN and $\mathcal{F}_{\theta}$. From Table \textcolor{red}{\ref{tab:Ablation-1}}, each metric has been improved after adding the proposed MSDN. These metrics show that it is feasible to obtain spatial details generated from the memory-based network. These metrics are further improved by adding NIN to our framework, which demonstrates the significance of NIN.
 % All of these results show the effectiveness of our method.

 \noindent\textbf{Size of the memory bank.} We also conduct experiments to demonstrate the impact of the $\mathbf N$ size of the memory bank in the MSDN. In the experiments, we modify $\mathbf{N}$ to $\mathbf{N}=\{16, 32, 64, 128\}$. The results are shown in Table \textcolor{red}{\ref{tab:Ablation-4}}. We observe that better pan-sharpening performance can be achieved by increasing the sizes from 16 to 64. However, performance gets worse as the size increases to 128.

  \noindent\textbf{Depth of NIN.} To explore the impact of depth of NIN, we experiment with various depths. In Table \textcolor{red}{\ref{tab:Ablation-3}}, we can see that better pan-sharpening performance can be achieved by increasing the layers from 1 to 3. However, continuing to increase the depth can not improve the performance but add computational complexity. Therefore, in our implementation, we choose the 3 depths of NIN. 

\begin{figure*}[!t]
    \centering
    \includegraphics[width=0.9\linewidth]{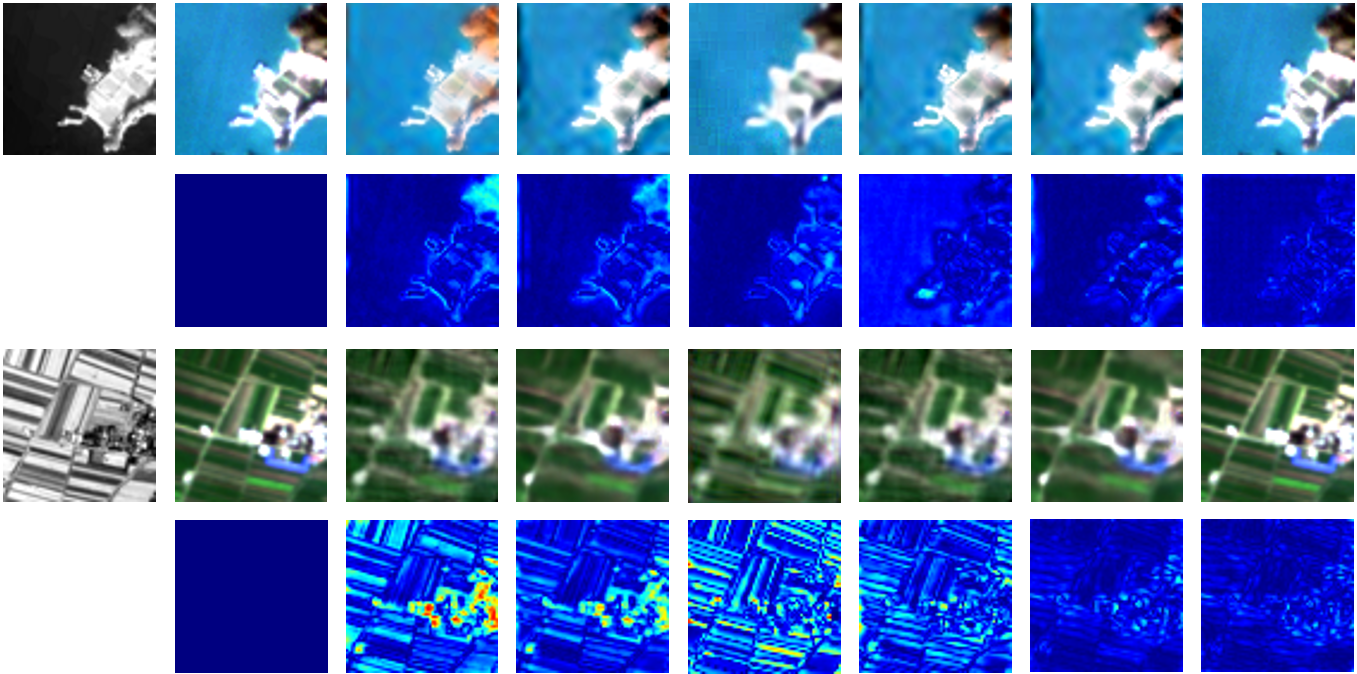}
   \caption{Visual results generated by different typical pan-sharpening methods. From left to right: PAN, Ground-Truth (GT), Brovey, MTF-GLP-HPM, PANNet, FusionNet, ADKNet and Ours. MAE denotes the (normalized) mean absolute error across all spectral bands. }
	\label{fig:visual result}
\end{figure*}

\subsection{Visualization of Spatial Details}
 
To illustrate the effectiveness of our proposed MSDN, we visualize the spatial details generated $\mathbf{P_s}$ in different scenes including urban areas, green vegetation, and water scenarios. As shown in Figure \textcolor{red}{\ref{fig:DRN effect}}, we can find that the generated $\mathbf{P_s}$ is close to the $\mathbf{HP}$ extracted from the PAN images. From the second and third columns in Figure \textcolor{red}{\ref{fig:DRN effect}}, although the scene is complex, MSDN can still generate spatial details similar to $\mathbf{HP}$. All results demonstrate that MSDN has great generation ability.

\subsection{Discussion of Multiscale Outputs}  
As mentioned in subsection \ref{framework}, our MSDN can generate different upscaled sizes of spatial details $\mathbf{P_s}$. To illustrate its versatility, we stack the MSDN and extend the framework to the multiscale framework. Thus, different upscaled sizes of HRMS can also be predicted. From Table \textcolor{red}{\ref{tab:Ablation-2}} and Figure \textcolor{red}{\ref{fig:Multiscale visual}}, we can see that the multiscale framework further improves all the metrics and the different upscaled outputs are close to their corresponding GT images. The above results verify the versatility of our MSDN. 

\subsection{Pan-sharpening on reduced-resolution scene}
 
\begin{table*}[!t]\normalsize

\renewcommand{\arraystretch}{1.3}
\begin{center}
\caption{Quantitative comparison on the GaoFen-1 and WorldView-4 dataset at reduced-resolution. Values are kept to three decimal places. The up- or down-arrow indicates that higher or lower metric corresponds to better images. The best results are highlighted in \textbf{bold}.}
\vspace{0.3cm}
\label{tab1: experiments result}
\setlength{\tabcolsep}{3mm}
\begin{tabular}{c|cccc|cccc}
\hline
\multirow{2}{*}{\textbf{Methods}} & \multicolumn{4}{c|}{\textbf{GaoFen-1}}   & \multicolumn{4}{c}{\textbf{WorldView-4}} \\ \cline{2-9} 
                         & \textbf{SAM}$\downarrow$   & \textbf{ERGAS}$\downarrow$  & \textbf{SCC}$\uparrow$   & \textbf{Q4}$\uparrow$    & \textbf{SAM}$\downarrow$    & \textbf{ERGAS}$\downarrow$  & \textbf{SCC}$\uparrow$   & \textbf{Q4}$\uparrow$    \\ \hline
GS \cite{GS_laben2000process_CS} & 0.213 & 13.933 & 0.732 & 0.512 & 0.213  & 9.379  & 0.682 & 0.521 \\
GSA \cite{GSA_aiazzi2007improving_CS}                     & 0.187 & 10.534 & 0.825 & 0.539 & 0.210  & 7.878  & 0.685 & 0.515 \\
Brovey \cite{Brovey_gillespie1987color_CS}                  & 0.183 & 14.041 & 0.764 & 0.567 & 0.206  & 8.689  & 0.706 & 0.544 \\
GFPCA \cite{8075405_GFPCA}                  & 0.182 & 11.163 & 0.819 & 0.571 & 0.205  & 7.807  & 0.725 & 0.563 \\
SFIM \cite{SFIM_liu2000smoothing_MRA}                 & 0.181 & 13.281 & 0.813 & 0.610 & 0.198  & 7.623  & 0.736 & 0.569 \\
MTF-GLP \cite{GLP_aiazzi2002context_MRA}                & 0.191 & 11.964 & 0.818 & 0.623 & 0.187  & 6.960  & 0.753 & 0.588 \\
MTF-GLP-HPM \cite{MTF-GLP-HPM-lee2009fast}           & 0.192 & 10.276 & 0.782 & 0.636 & 0.189  & 7.063  & 0.778 & 0.597 \\ \hline
PNN \cite{Masi2016PNN}                    & 0.144 & 8.509  & 0.875 & 0.640 & 0.173  & 6.495  & 0.843 & 0.628 \\
PANNet \cite{yang2017pannet}                  & 0.137 & 8.569  & 0.880 & 0.647 & 0.166  & 6.827  & 0.845 & 0.644 \\
MSDCNN \cite{Yuan2018AMSDCNN}                 & 0.135 & 8.215  & 0.883 & 0.657 & 0.156  & 6.606  & 0.845 & 0.648 \\
FusionNet \cite{deng2021detailFusionNet}              & 0.133 & 8.179  & 0.890 & 0.665 & 0.176  & 6.198  & 0.853 & {0.668} \\
ADKNet \cite{Peng2022ADKNet}                & {0.119} & {7.670}  & {0.894} & {0.674} & {0.144}  & {5.827}  & {0.877} & \textbf{{0.673}} \\ \hline
Ours                 & \textbf{0.099} &	\textbf{6.320} &	\textbf{0.912} &	\textbf{0.705}  & \textbf{{0.125}}  & \textbf{{5.681}}  & \textbf{{0.879}} & 0.654 \\ \hline
\end{tabular}
\end{center}
\vspace{-0.5cm}
\end{table*}

 \begin{table}[!t] \normalsize
    \begin{center}
    \caption{Quantitative comparison on the WorldView-4 datase with MS image super-resolution methods. The best results are highlighted in \textbf{bold}. }
    \label{tab:MS-Superresolution}
    \renewcommand{\arraystretch}{1.3}
    \setlength{\tabcolsep}{2mm}
    \begin{tabular}{c|cccc}
        \hline
          \textbf{Methods}          & \textbf{SAM}$\downarrow$& \textbf{ERGAS}$\downarrow$& \textbf{SCC}$\uparrow$ & \textbf{Q4}$\uparrow$ \\  \hline

        SRCNN  \cite{dong2015image}    & 0.140   & 6.909 & 0.824 & 0.531 \\
        VDSR  \cite{kim2016accurate}    &0.144    &6.883  & 0.828 & 0.539  \\
        DCM  \cite{haut2019remote}     & 0.148   & 7.139 & 0.816	& 0.515   \\
        LGCNet \cite{lei2017super}   & 0.142   & 7.018	& 0.823 & 0.525  \\
        HSENet \cite{lei2021hybrid}   &0.136    &6.397  & 0.845 & 0.564   \\ \hline
        Ours &\textbf{ 0.125}&\textbf{5.681}&\textbf{0.879}& \textbf{0.654}    \\ \hline
    \end{tabular}
    \end{center}

    \vspace{-0.3cm}
\end{table}

\begin{figure}[!t]
	\centering
	
    %第一行图片展示
		%左标题1
		
    \rotatebox{90}{\scriptsize{~~~~\textbf{RGB}}}
    \begin{minipage}[t]{0.127\linewidth}
			\centering
			\includegraphics[width=1\linewidth]{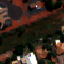}
	\end{minipage}
 	\begin{minipage}[t]{0.127\linewidth}
			\centering
			\includegraphics[width=1\linewidth]{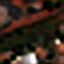}
	\end{minipage}
 	\begin{minipage}[t]{0.127\linewidth}
			\centering
			\includegraphics[width=1\linewidth]{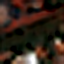}
	\end{minipage}
 	\begin{minipage}[t]{0.127\linewidth}
			\centering
			\includegraphics[width=1\linewidth]{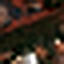}
	\end{minipage}
 	\begin{minipage}[t]{0.127\linewidth}
			\centering
			\includegraphics[width=1\linewidth]{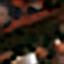}
	\end{minipage}
 	\begin{minipage}[t]{0.127\linewidth}
			\centering
			\includegraphics[width=1\linewidth]{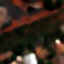}
	\end{minipage}
 	\begin{minipage}[t]{0.127\linewidth}
			\centering
			\includegraphics[width=1\linewidth]{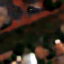}
	\end{minipage}
 
    % 两行图片的间隙有点大，通过vspace进行微调
	% \vspace{1mm}

    % 第二行图片展示
        % 左标题2
        
    \rotatebox{90}{\scriptsize{~~~~\textbf{MAE}}}
 	\begin{minipage}[t]{0.127\linewidth}
			\centering
			\includegraphics[width=1\linewidth]{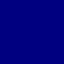}
	\end{minipage}
 	\begin{minipage}[t]{0.127\linewidth}
			\centering
			\includegraphics[width=1\linewidth]{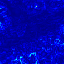}
	\end{minipage}
 	\begin{minipage}[t]{0.127\linewidth}
			\centering
			\includegraphics[width=1\linewidth]{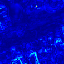}
	\end{minipage}
 	\begin{minipage}[t]{0.127\linewidth}
			\centering
			\includegraphics[width=1\linewidth]{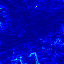}
	\end{minipage}
 	\begin{minipage}[t]{0.127\linewidth}
			\centering
			\includegraphics[width=1\linewidth]{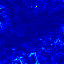}
	\end{minipage}
 	\begin{minipage}[t]{0.127\linewidth}
			\centering
			\includegraphics[width=1\linewidth]{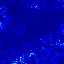}
	\end{minipage}
        \begin{minipage}[t]{0.127\linewidth}
			\centering
			\includegraphics[width=1\linewidth]{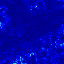}
	\end{minipage}
	
      \vspace{0.7mm}
      
%%%%%%%%%%%%%%%%%%%%%%%%%%%%%%%%%%%%%%%%%%%%%第二幅
    %第一行图片展示
		%左标题1
		
    \rotatebox{90}{\scriptsize{~~~~\textbf{RGB}}}
    \begin{minipage}[t]{0.127\linewidth}
			\centering
			\includegraphics[width=1\linewidth]{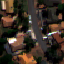}
	\end{minipage}
 	\begin{minipage}[t]{0.127\linewidth}
			\centering
			\includegraphics[width=1\linewidth]{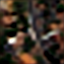}
	\end{minipage}
 	\begin{minipage}[t]{0.127\linewidth}
			\centering
			\includegraphics[width=1\linewidth]{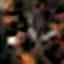}
	\end{minipage}
 	\begin{minipage}[t]{0.127\linewidth}
			\centering
			\includegraphics[width=1\linewidth]{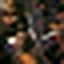}
	\end{minipage}
 	\begin{minipage}[t]{0.127\linewidth}
			\centering
			\includegraphics[width=1\linewidth]{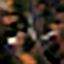}
	\end{minipage}
 	\begin{minipage}[t]{0.127\linewidth}
			\centering
			\includegraphics[width=1\linewidth]{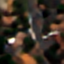}
	\end{minipage}
 	\begin{minipage}[t]{0.127\linewidth}
			\centering
			\includegraphics[width=1\linewidth]{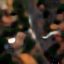}
	\end{minipage}
 
    % 两行图片的间隙有点大，通过vspace进行微调
	% \vspace{1mm}

    % 第二行图片展示
        % 左标题2
        
    \rotatebox{90}{\scriptsize{~~~~\textbf{MAE}}}
 	\begin{minipage}[t]{0.127\linewidth}
			\centering
			\includegraphics[width=1\linewidth]{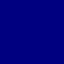}
	\end{minipage}
 	\begin{minipage}[t]{0.127\linewidth}
			\centering
			\includegraphics[width=1\linewidth]{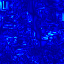}
	\end{minipage}
 	\begin{minipage}[t]{0.127\linewidth}
			\centering
			\includegraphics[width=1\linewidth]{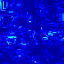}
	\end{minipage}
 	\begin{minipage}[t]{0.127\linewidth}
			\centering
			\includegraphics[width=1\linewidth]{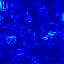}
	\end{minipage}
 	\begin{minipage}[t]{0.127\linewidth}
			\centering
			\includegraphics[width=1\linewidth]{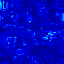}
	\end{minipage}
 	\begin{minipage}[t]{0.127\linewidth}
			\centering
			\includegraphics[width=1\linewidth]{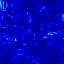}
	\end{minipage}
        \begin{minipage}[t]{0.127\linewidth}
			\centering
			\includegraphics[width=1\linewidth]{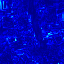}
	\end{minipage}
	
      \vspace{0.7mm}

%%%%%%%%%%%%%%%%%%%%%%%%%%%%%%%%%%%%%%%%%%%%%第三幅  
		%左标题1
		
    \rotatebox{90}{\scriptsize{~~~~\textbf{RGB}}}
    \begin{minipage}[t]{0.127\linewidth}
			\centering
			\includegraphics[width=1\linewidth]{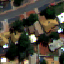}
	\end{minipage}
 	\begin{minipage}[t]{0.127\linewidth}
			\centering
			\includegraphics[width=1\linewidth]{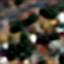}
	\end{minipage}
 	\begin{minipage}[t]{0.127\linewidth}
			\centering
			\includegraphics[width=1\linewidth]{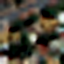}
	\end{minipage}
 	\begin{minipage}[t]{0.127\linewidth}
			\centering
			\includegraphics[width=1\linewidth]{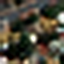}
	\end{minipage}
 	\begin{minipage}[t]{0.127\linewidth}
			\centering
			\includegraphics[width=1\linewidth]{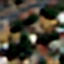}
	\end{minipage}
 	\begin{minipage}[t]{0.127\linewidth}
			\centering
			\includegraphics[width=1\linewidth]{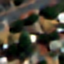}
	\end{minipage}
 	\begin{minipage}[t]{0.127\linewidth}
			\centering
			\includegraphics[width=1\linewidth]{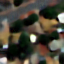}
	\end{minipage}
 
    % 两行图片的间隙有点大，通过vspace进行微调
	% \vspace{1mm}

    % 第二行图片展示
        % 左标题2
        
    \rotatebox{90}{\scriptsize{~~~~\textbf{MAE}}}
 	\begin{minipage}[t]{0.127\linewidth}
			\centering
			\includegraphics[width=1\linewidth]{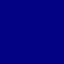}
	\end{minipage}
 	\begin{minipage}[t]{0.127\linewidth}
			\centering
			\includegraphics[width=1\linewidth]{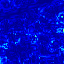}
	\end{minipage}
 	\begin{minipage}[t]{0.127\linewidth}
			\centering
			\includegraphics[width=1\linewidth]{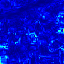}
	\end{minipage}
 	\begin{minipage}[t]{0.127\linewidth}
			\centering
			\includegraphics[width=1\linewidth]{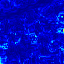}
	\end{minipage}
 	\begin{minipage}[t]{0.127\linewidth}
			\centering
			\includegraphics[width=1\linewidth]{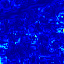}
	\end{minipage}
 	\begin{minipage}[t]{0.127\linewidth}
			\centering
			\includegraphics[width=1\linewidth]{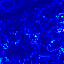}
	\end{minipage}
        \begin{minipage}[t]{0.127\linewidth}
			\centering
			\includegraphics[width=1\linewidth]{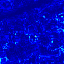}
	\end{minipage}
	
      \vspace{0.7mm}

%%%%%%%%%%%%%%%%%%%%%%%%%%%%%%%%%%%%%%%%%%%%%第四幅  
		%左标题1
     \rotatebox{90}{\scriptsize{~~~~\textbf{RGB}}}
        \begin{minipage}[t]{0.127\linewidth}
    			\centering
    			\includegraphics[width=1\linewidth]{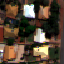}
        \end{minipage}
        \begin{minipage}[t]{0.127\linewidth}
                \centering
                \includegraphics[width=1\linewidth]{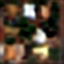}
        \end{minipage}
        \begin{minipage}[t]{0.127\linewidth}
                \centering
                \includegraphics[width=1\linewidth]{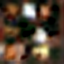}
        \end{minipage}
        \begin{minipage}[t]{0.127\linewidth}
                \centering
                \includegraphics[width=1\linewidth]{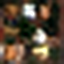}
        \end{minipage}
        \begin{minipage}[t]{0.127\linewidth}
                \centering
                \includegraphics[width=1\linewidth]{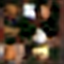}
        \end{minipage}
        \begin{minipage}[t]{0.127\linewidth}
                \centering
                \includegraphics[width=1\linewidth]{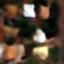}
        \end{minipage}
        \begin{minipage}[t]{0.127\linewidth}
                \centering
                \includegraphics[width=1\linewidth]{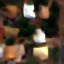}
        \end{minipage}
     
        % 两行图片的间隙有点大，通过vspace进行微调
        % \vspace{1mm}
    
        % 第二行图片展示
            % 左标题2

        \rotatebox{90}{\scriptsize{~~~~\textbf{MAE}}}
        \begin{minipage}[t]{0.127\linewidth}
                \centering
                \includegraphics[width=1\linewidth]{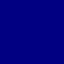}
        \end{minipage}
        \begin{minipage}[t]{0.127\linewidth}
                \centering
                \includegraphics[width=1\linewidth]{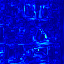}
        \end{minipage}
        \begin{minipage}[t]{0.127\linewidth}
                \centering
                \includegraphics[width=1\linewidth]{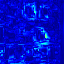}
        \end{minipage}
        \begin{minipage}[t]{0.127\linewidth}
                \centering
                \includegraphics[width=1\linewidth]{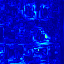}
        \end{minipage}
        \begin{minipage}[t]{0.127\linewidth}
                \centering
                \includegraphics[width=1\linewidth]{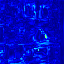}
        \end{minipage}
        \begin{minipage}[t]{0.127\linewidth}
                \centering
                \includegraphics[width=1\linewidth]{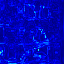}
        \end{minipage}
            \begin{minipage}[t]{0.127\linewidth}
                \centering
                \includegraphics[width=1\linewidth]{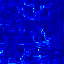}
        \end{minipage}
        
      \vspace{0.7mm}
 
	\caption{Visualization results generated by different super-resolution methods (left to right: Ground-Truth (GT), SRCNN \cite{dong2015image}, VDSR \cite{kim2016accurate}, DCM \cite{haut2019remote}, LGCNet \cite{lei2017super}, HSENet \cite{lei2021hybrid} and Ours). MAE denotes the (normalized) mean absolute error across all spectral bands. }

	\label{fig:super-visual_result}
\end{figure}

 \noindent\textbf{Quantitative comparison.}
The quantitative comparison results over Gaofen1 and WorldView-4 satellite datasets are reported in Table~\ref{tab1: experiments result}, where the best results are highlighted in bold. As shown in Table \textcolor{red}{\ref{tab1: experiments result}}, although our method performs pan-sharpening without PAN image in the testing phase, the performance of our framework still surpasses the other promising pan-sharpening methods. Furthermore, through the SAM and ERGAS metrics, we can observe that our results preserve more spectral information due to the absence of PAN image features. Meanwhile, our method also achieves competitive results on spatial metric SCC. The above results well prove the effectiveness of our method, which only uses MS image for pan-sharpening while ensuring the sharping performance.

\begin{figure*}[!tbp]
	\centering

    %第一行图片展示
		%左标题1
	\begin{minipage}[t]{0.13\linewidth}
		\centering
		  \includegraphics[width=1\linewidth]{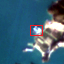}
	\end{minipage}
        \begin{minipage}[t]{0.13\linewidth}
		\centering
		\includegraphics[width=1\linewidth]{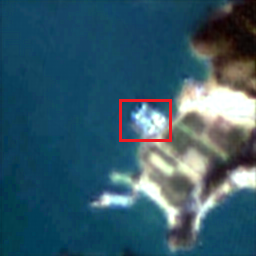}
	\end{minipage}\hspace{0.05mm}
 	\begin{minipage}[t]{0.13\linewidth}
		\centering
		\includegraphics[width=1\linewidth]{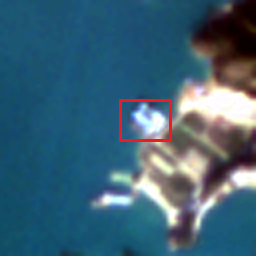}
	\end{minipage}\hspace{0.05mm}
 	\begin{minipage}[t]{0.13\linewidth}
		\centering
		\includegraphics[width=1\linewidth]{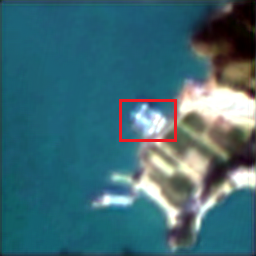}
	\end{minipage}\hspace{0.05mm}
 	\begin{minipage}[t]{0.13\linewidth}
		\centering
		\includegraphics[width=1\linewidth]{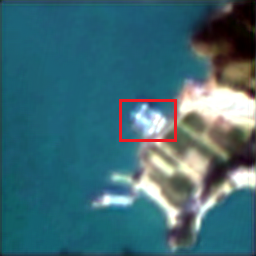}
	\end{minipage}\hspace{0.05mm}
 	\begin{minipage}[t]{0.13\linewidth}
		\centering
		\includegraphics[width=1\linewidth]{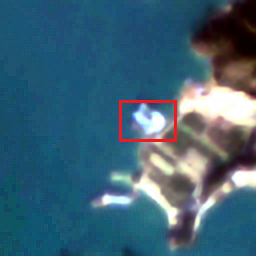}
	\end{minipage}\hspace{0.05mm}
        \begin{minipage}[t]{0.13\linewidth}
		\centering
		\includegraphics[width=1\linewidth]{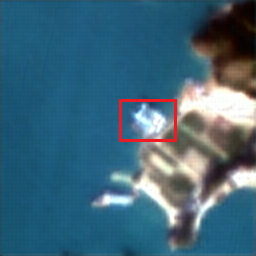}
	\end{minipage}
 
    % 两行图片的间隙有点大，通过vspace进行微调
	% \vspace{1mm}

    \vspace{0.12cm}
      
%%%%%%%%%%%%%%%%%%%%%%%%%%%%%%%%%%%%%%%%%%%%%
        \begin{minipage}[t]{0.13\linewidth}
		\centering
		\includegraphics[width=1\linewidth]{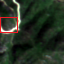}
	\end{minipage}
        \begin{minipage}[t]{0.13\linewidth}
		\centering
		\includegraphics[width=1\linewidth]{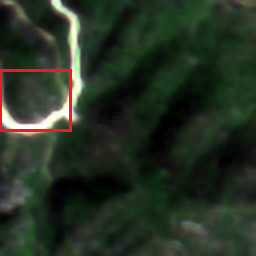}
	\end{minipage}\hspace{0.05mm}
 	\begin{minipage}[t]{0.13\linewidth}
		\centering
		\includegraphics[width=1\linewidth]{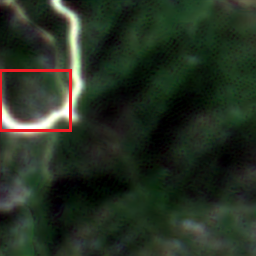}
	\end{minipage}\hspace{0.05mm}
 	\begin{minipage}[t]{0.13\linewidth}
		\centering
		\includegraphics[width=1\linewidth]{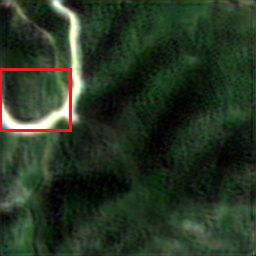}
	\end{minipage}\hspace{0.05mm}
 	\begin{minipage}[t]{0.13\linewidth}
		\centering
		\includegraphics[width=1\linewidth]{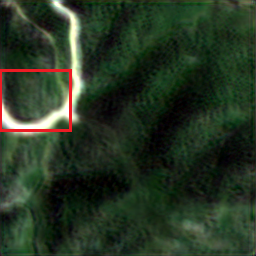}
	\end{minipage}\hspace{0.05mm}
 	\begin{minipage}[t]{0.13\linewidth}
		\centering
		\includegraphics[width=1\linewidth]{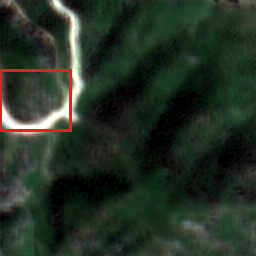}
	\end{minipage}\hspace{0.05mm}
        \begin{minipage}[t]{0.13\linewidth}
		\centering
		\includegraphics[width=1\linewidth]{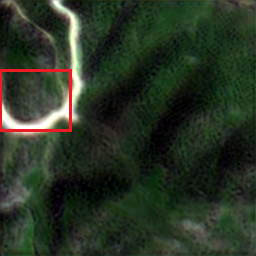}
	\end{minipage}
 
    % 两行图片的间隙有点大，通过vspace进行微调
	% \vspace{1mm}

    \vspace{0.12cm}
      
%%%%%%%%%%%%%%%%%%%%%%%%%%%%%%%%%%%%%%%%%%%%%
        \begin{minipage}[t]{0.13\linewidth}
		\centering
		\includegraphics[width=1\linewidth]{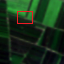}
	\end{minipage}
        \begin{minipage}[t]{0.13\linewidth}
		\centering
		\includegraphics[width=1\linewidth]{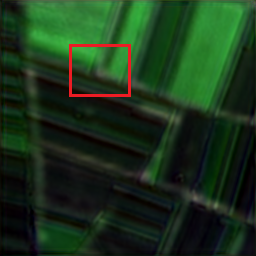}
	\end{minipage}\hspace{0.05mm}
 	\begin{minipage}[t]{0.13\linewidth}
		\centering
		\includegraphics[width=1\linewidth]{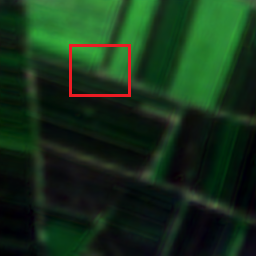}
	\end{minipage}\hspace{0.05mm}
 	\begin{minipage}[t]{0.13\linewidth}
		\centering
		\includegraphics[width=1\linewidth]{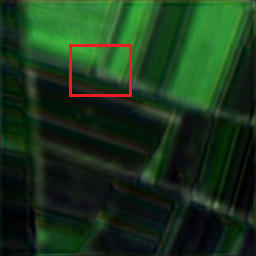}
	\end{minipage}\hspace{0.05mm}
 	\begin{minipage}[t]{0.13\linewidth}
		\centering
		\includegraphics[width=1\linewidth]{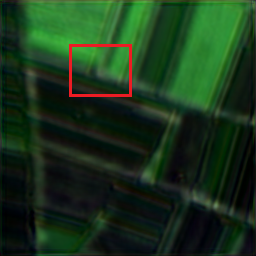}
	\end{minipage}\hspace{0.05mm}
 	\begin{minipage}[t]{0.13\linewidth}
		\centering
		\includegraphics[width=1\linewidth]{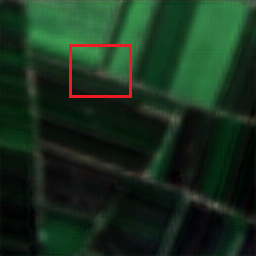}
	\end{minipage}\hspace{0.05mm}
        \begin{minipage}[t]{0.13\linewidth}
		\centering
		\includegraphics[width=1\linewidth]{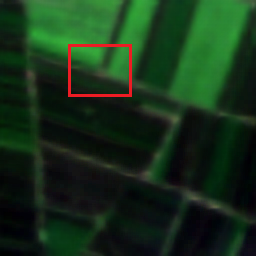}
	\end{minipage}
 
    % 两行图片的间隙有点大，通过vspace进行微调
	% \vspace{1mm}

    \vspace{0.12cm}
      
%%%%%%%%%%%%%%%%%%%%%%%%%%%%%%%%%%%%%%%%%%%%%
        \begin{minipage}[t]{0.13\linewidth}
		\centering
		\includegraphics[width=1\linewidth]{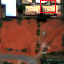}
	\end{minipage}
        \begin{minipage}[t]{0.13\linewidth}
		\centering
		\includegraphics[width=1\linewidth]{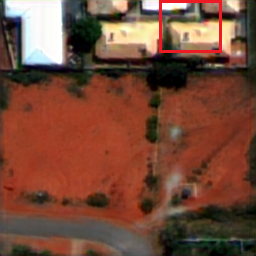}
	\end{minipage}\hspace{0.05mm}
 	\begin{minipage}[t]{0.13\linewidth}
		\centering
	    \includegraphics[width=1\linewidth]{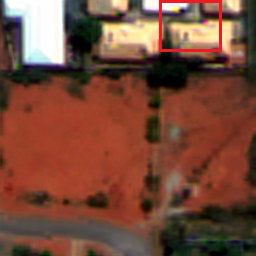}
	\end{minipage}\hspace{0.05mm}
 	\begin{minipage}[t]{0.13\linewidth}
		\centering
		\includegraphics[width=1\linewidth]{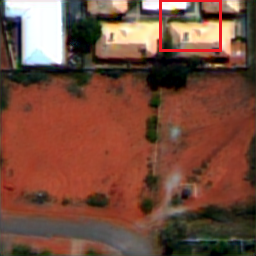}
	\end{minipage}\hspace{0.05mm}
 	\begin{minipage}[t]{0.13\linewidth}
		\centering
		\includegraphics[width=1\linewidth]{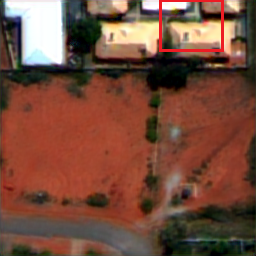}
	\end{minipage}\hspace{0.05mm}
 	\begin{minipage}[t]{0.13\linewidth}
		\centering
		\includegraphics[width=1\linewidth]{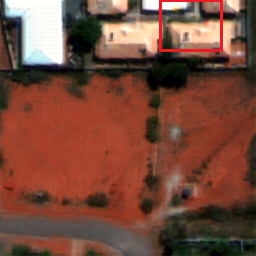}
	\end{minipage}\hspace{0.05mm}
        \begin{minipage}[t]{0.13\linewidth}
		\centering
		\includegraphics[width=1\linewidth]{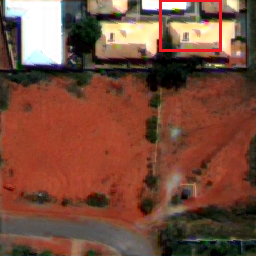}
	\end{minipage}
 
    % 两行图片的间隙有点大，通过vspace进行微调
	% \vspace{1mm}

    \vspace{0.12cm}

%%%%%%%%%%%%%%%%%%%%%%%%%%%%%%%%%%%%%%%%%%%%%
        \begin{minipage}[t]{0.13\linewidth}
		\centering
		\includegraphics[width=1\linewidth]{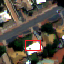}
	\end{minipage}
        \begin{minipage}[t]{0.13\linewidth}
		\centering
		\includegraphics[width=1\linewidth]{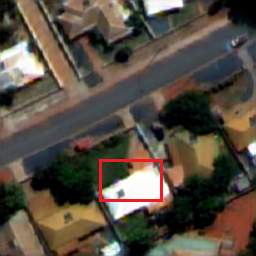}
	\end{minipage}\hspace{0.05mm}
 	\begin{minipage}[t]{0.13\linewidth}
		\centering
		\includegraphics[width=1\linewidth]{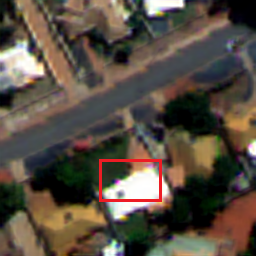}
	\end{minipage}\hspace{0.05mm}
 	\begin{minipage}[t]{0.13\linewidth}
		\centering
		\includegraphics[width=1\linewidth]{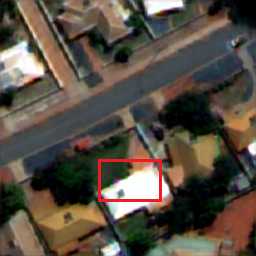}
	\end{minipage}\hspace{0.05mm}
 	\begin{minipage}[t]{0.13\linewidth}
		\centering
		\includegraphics[width=1\linewidth]{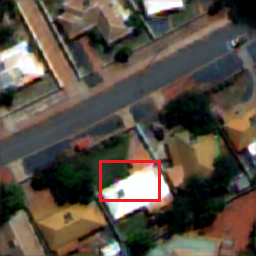}
	\end{minipage}\hspace{0.05mm}
 	\begin{minipage}[t]{0.13\linewidth}
		\centering
		\includegraphics[width=1\linewidth]{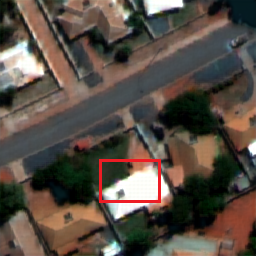}
	\end{minipage}\hspace{0.05mm}
        \begin{minipage}[t]{0.13\linewidth}
		\centering
		\includegraphics[width=1\linewidth]{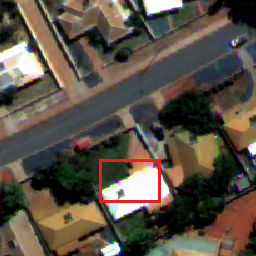}
	\end{minipage}
 
    % 两行图片的间隙有点大，通过vspace进行微调
	% \vspace{1mm}

    \vspace{0.4cm}
%%%%%%%%%%%%%%%%%%%%%%%%%%%%%%%%%%%%%%%%%%%%%
   
    \caption{Full-resolution visualization results generated by different DL-based pan-sharpening methods (left to right: MS image, PNN \cite{Masi2016PNN}, PANNet \cite{yang2017pannet}, MSDCNN \cite{Yuan2018AMSDCNN}, FusionNet \cite{deng2021detailFusionNet}, ADKNet \cite{Peng2022ADKNet} and Ours). The first three rows are sampled from the Gaofen1 dataset, while the last two rows are from the WorldView-4 dataset.}
    
    \label{fig:full_visual result}

\end{figure*}

 \noindent\textbf{Qualitative comparison.}
We also show the qualitative comparison of the visual results to prove the effectiveness of our method. We randomly select images from the Gaofen1 and WordView-4 datasets to present the RGB channel of output results as shown in Figure \textcolor{red}{\ref{fig:visual result}}. The mean absolute error (MAE) is used to calculate the difference between the outputs and the ground truth. Compared with the other pan-sharpening methods, our method has a higher coincidence with the ground truth, which demonstrates that our model has minor spatial and spectral distortions. In general, these visualisation results further testify to the superiority of MSDN in spectral and spatial fidelity.

\begin{table}[!t]\normalsize
\begin{center}
    \caption{The non-reference metrics on the WorldView-4 dataset at full-resolution. Best results highlighted in \textbf{bold}.}
    \label{tab:full-resolution}
    \renewcommand{\arraystretch}{1.3}
    \setlength{\tabcolsep}{3.5mm}
\begin{tabular}{c|lll}
\hline
\textbf{Methods}   & \multicolumn{1}{c}{\textbf{QNR}$\uparrow$} & \multicolumn{1}{c}{$\mathbf{D_s}$$\downarrow$} & \multicolumn{1}{c}{$\mathbf{D_\lambda}$$\downarrow$} \\ \hline
PNN \cite{Masi2016PNN} & 0.736                    & 0.161                  & 0.131  \\
PANNet \cite{yang2017pannet}     & 0.739                   & 0.159                  & 0.129                  \\
MSDCNN \cite{Yuan2018AMSDCNN}   & 0.745                   & 0.174                  & 0.122                  \\
FusionNet \cite{deng2021detailFusionNet} & 0.759                   & 0.128                  & 0.128                  \\
ADKNet \cite{Peng2022ADKNet} & 0.757              & 0.145                  & 0.121                  \\   \hline
Ours   & \textbf{0.812}          & \textbf{0.115}         & \textbf{0.086}         \\ \hline
\end{tabular}
\end{center}

\vspace{-0.5cm}
\end{table}

\subsection{Comparisons with MS image super-resolution methods}
 \noindent\textbf{Quantitative comparison.} Since our method only requires the MS image in the inference phase like the super-resolution method, we additionally conduct comparison experiments with other promising super-resolution methods of the MS image to illustrate the effectiveness of the method from the side. As shown in Table \textcolor{red}{\ref{tab:MS-Superresolution}}, obviously, despite performing well on spectral metrics SAM and ERGAS, super-resolution methods are inferior to pan-sharpening methods on the spatial metric SCC due to the lack of PAN images. However, our method achieves 0.879 on SCC while also obtaining the highest score on the spectral metrics, which verifies its effectiveness in improving the spatial resolution of super-resolution methods.

 \noindent\textbf{Qualitative comparison.} We also visualize super-resolution images for comparison with other methods. As shown in Figure~\textcolor{red}{\ref{fig:super-visual_result}}, we can see from the MAE images that the super-resolution images obtained from our method are the closest to the reference image, which verifies the quantitative results on the super-resolution task.

\subsection{Pan-sharpening on full-resolution scene}
 \noindent\textbf{Quantitative comparison.} In order to demonstrate the real-world application value, we also conduct experiments on full-resolution scenes obtained from the WorldView-4 satellite. A quantitative comparison between promising DL-based methods and our framework is presented in the following Table \textcolor{red}{\ref{tab:full-resolution}}. The lower $\mathrm D_\lambda$, $\mathrm D_s$, and the higher QNR correspond to the better image quality, where the best results are highlighted in bold. As can be seen obviously, our methods overpass other competitive pan-sharpening methods, which demonstrate excellent ability in real-world application.

 \noindent\textbf{Qualitative comparison.} We further perform a full-resolution visualization on the Gaofen1 and WorldView-4 datasets. The qualitative results are shown in Figure~\textcolor{red}{\ref{fig:full_visual result}}. From the red box, we can observe that our method outperforms other methods in both spectral fidelity and spatial resolution. Especially, from the third row of Figure~\textcolor{red}{\ref{fig:full_visual result}}, we also find that our method can well solve the artefacts and texture blurring problem caused by the misalignment between PAN and MS images, which benefits from the fact that our method does not require PAN images in practical applications. The above results demonstrate the effectiveness and universality of our method.

% ###############################################

%-------------------------------------------------------------------------
\section{Conclusion} \label{conclusion}
In this work, we formulated a new representation of the spatial details and proposed a novel memory-based network called MSDN to generate the required spatial details instead of obtaining them from PAN images. To model it, we decomposed the MSDN into two subnetworks: a memory-controlled subnetwork and a weighted coefficient subnetwork. In the memory-controlled subnetwork, we encoded the MS features in the memory encoder to query the corresponding spatial details from the memory bank and then obtained the spatial features through the memory decoder. As for the weighted coefficient subnetwork, we utilized channel attention to predict the weighted features. The two subnetworks are tightly coupled and generate the required spatial details. We also conducted an MSDN-based pan-sharpening framework to evaluate its effectiveness. The framework was easy to extend and can be trained in an end-to-end manner. The results of additional ablation studies support the effectiveness of the proposed components. In addition, our method outperforms other conventional and state-of-the-art methods in both reduced- and full-resolution experiments on GF1 and WorldView-4 datasets, indicating the good generalisation of the proposed method. For future work, we will explore the method of utilizing PAN images to directly generate pan-sharpening images and embed it into our proposed framework. We believe our method will be useful in real-world applications.

\bibliographystyle{IEEEtran}
\bibliography{reference.bib}

\vfill

\end{document}